\pdfoutput=1

\documentclass[11pt]{article}

\usepackage[preprint]{acl}

\usepackage{times}
\usepackage{latexsym}

\usepackage[T1]{fontenc}

\usepackage[utf8]{inputenc}

\usepackage{microtype}

\usepackage{inconsolata}

\usepackage{graphicx}

\usepackage{multirow}
\usepackage{colortbl}
\usepackage{listings}
\usepackage{float}
\usepackage{lineno}
\usepackage{csquotes}
\usepackage{makecell}

\usepackage{graphicx}
\usepackage{amsmath}
\usepackage{amssymb}
\usepackage{booktabs}
\usepackage{multirow} 
\usepackage{float} 
\usepackage{multicol}
\usepackage{pifont}%

\newcommand*\colourcheck[1]{%
  \expandafter\newcommand\csname #1check\endcsname{\textcolor{#1}{\ding{51}}}%
}
\colourcheck{green}
\newcommand*\colourx[1]{%
  \expandafter\newcommand\csname #1x\endcsname{\textcolor{#1}{\ding{55}}}%
}
\colourx{red}

\newcommand{\cmark}{\greencheck}%
\newcommand{\xmark}{\redx}%

\definecolor{codegreen}{rgb}{0,0.6,0}
\definecolor{codegray}{rgb}{0.5,0.5,0.5}
\definecolor{codepurple}{rgb}{0.58,0,0.82}
\definecolor{backcolour}{rgb}{0.95,0.95,0.92}

\lstdefinestyle{coloredpython}{
  backgroundcolor=\color{backcolour}, commentstyle=\color{codegreen},
  keywordstyle=\color{magenta},
  numberstyle=\tiny\color{codegray},
  stringstyle=\color{codepurple},
  basicstyle=\ttfamily\footnotesize,
  breakatwhitespace=false,         
  breaklines=true,                 
  captionpos=b,                    
  keepspaces=true,                 
  numbers=left,                    
  numbersep=5pt,                  
  showspaces=false,                
  showstringspaces=false,
  showtabs=false,                  
  tabsize=2
}
\newcommand{\benchmarknum}{6 }

\newcommand{\dsnum}{31 }
\newcommand{\modelname}{\textsc{LATTE}}
\newcommand{\tool}[1]{{\textsc{#1}}}
\newcommand{\datastep}[1]{{\textbf{\textsc{#1}}}}
\newcommand{\cotht}{CoT}
\newcommand{\cota}{LATTE-trace}
\newcommand{\cotas}{LATTE-traces}
\newcommand{\da}{Direct} 
\newcommand{\eg}{e.\,g.\ }

\title{\includegraphics[height=5mm]{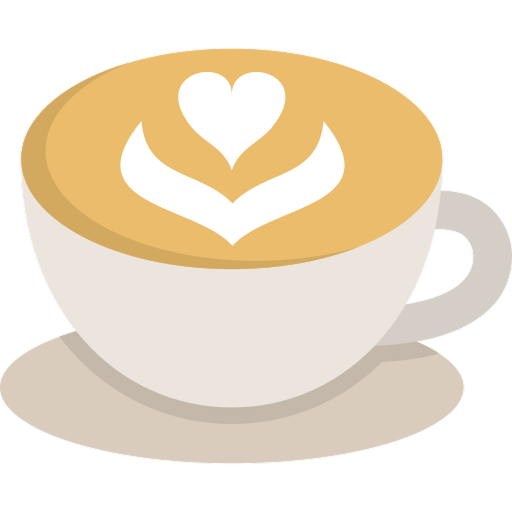} \modelname: Learning to Think with Vision Specialists}

\author{
\\
Zixian Ma$^{1,2}$\thanks{Work done while interning at Salesforce Research}, Jianguo Zhang$^2$, Zhiwei Liu$^2$, Jieyu Zhang$^1$, Juntao Tan$^2$, Manli Shu$^2$, \\ Juan Carlos Niebles$^2$, Shelby Heinecke$^2$, Huan Wang$^2$, Caiming Xiong$^2$, \\ Ranjay Krishna$^1$, Silvio Savarese$^2$ \\
$^1$University of Washington, $^2$Salesforce Research\\ \\
Website: \url{https://latte-web.github.io} \\
Code: \url{https://github.com/SalesforceAIResearch/LATTE} 
}

\begin{document}
\maketitle
\begin{abstract}
While open-source vision-language models perform well on simple question-answering, they still struggle with complex questions that require both perceptual and reasoning capabilities. We propose \modelname, a family of vision-language models that have LeArned to Think wiTh vision spEcialists. By offloading perception to state-of-the-art vision models, our approach enables vision-language models to focus solely on reasoning over high-quality perceptual information.  To train \modelname, we synthesize and filter a large dataset of 273K multi-modal reasoning traces over perceptual outputs of vision specialists. \modelname\ trained on this data achieves significant 4-5\% gains over baselines across \benchmarknum benchmarks covering both perception and reasoning abilities. Ablation studies reveal that the effectiveness of multi-modal reasoning traces depends on the data sources, formats, and quality of thoughts. 


\end{abstract} 

\begin{figure*}[ht]
    \centering
    \includegraphics[width=\textwidth]{images/teaser_small.pdf}
    \vspace{-5pt}
    \caption{\textbf{Example outputs of \modelname\ vs. SoTA open-source multi-modal language models.} Our \modelname\ model is able to answer challenging visual questions by reasoning over perceptual information output by vision specialists. It does so by generating a reasoning trace over vision specialists' outputs and producing a final answer.}
    \label{fig:teaser}
    \vspace{-10pt}
\end{figure*}
\section{Introduction}
\label{sec:intro}

The landscape of real-world vision-language tasks is vast, spanning from basic visual question answering~\cite{antol2015vqa}, fine-grained object recognition to complex multi-step geometric reasoning~\cite{hu2024visualsketchpadsketchingvisual}. 
These tasks demand both perception and reasoning. For instance, a user might photograph a gas price panel and ask how much fuel they can afford within a given budget (Figure~\ref{fig:teaser}). Solving this requires a model with strong perception—localizing prices via OCR—and multi-step reasoning to compute the answer. 
While proprietary models like GPT-4o excel due to extensive data and model size scaling, smaller open-source models still struggle~\cite{ma2024mms}.

To narrow the gap between large proprietary models and smaller open-source counterparts within a reasonable budget, researchers have explored distilling perception and reasoning from larger vision-language models~\cite{shao2024visualcot,xu2025llavacotletvisionlanguage} or specialized vision models~\cite{hu2024visual}.
Despite these efforts, open-source models continue to lag behind.

We argue that the primary reason for this lag is the perception limitations of small vision-language models. While open-source language models have largely caught up with their proprietary counterparts~\cite{lambert2024t,bi2024deepseek}, vision-language models have yet to master heterogeneous vision capabilities.
The computer vision community has historically tackled these capabilities separately—e.g., DepthAnything~\cite{depth_anything_v2} for depth estimation and GroundingDINO~\cite{liu2023grounding} for object recognition—while unified models still lag behind~\cite{lu2024unified}. Similarly, the human brain dedicates distinct regions to categorical recognition (ventral stream) and spatial reasoning (dorsal stream)~\cite{goodale1992separate}, with the reasoning and language-processing frontal and temporal lobes occupying a different volume~\cite{keller2012sensorimotor}.
By contrast, vision-language models remain heavily skewed toward language, treating visual encoders as an afterthought~\cite{deitke2024molmo}.

We depart from the \emph{learning to perceive and reason} paradigm to propose a new approach: \emph{learning to reason with vision specialists} (Figure \ref{fig:framework}). Rather than expecting a small model to master both perception and reasoning, we leverage decades of advancements in computer vision by relying on specialized vision models to provide perceptual information. This allows the vision-language model to focus exclusively on acquiring perceptual information from vision specialists and reasoning over them—enabling it to `\textit{see further by standing on the shoulders of giants}.'
Such a paradigm reduces the burden on models to extract low-level perceptual signals, allowing them to concentrate on higher-level reasoning.

 To implement this paradigm, we curate high-quality training data in the form of multi-step reasoning traces that integrate perceptual information from vision specialists. 
 We formulate the multi-step reasoning traces as \cota, where each step consists of: (1) a \textit{thought} for verbalized reasoning; (2) an \textit{action} to retrieve perceptual information from a specific vision specialist; and (3) an \textit{observation} of the returned data. 
Since obtaining these traces at scale with human annotators is costly, we develop two data engines for synthetic data generation.  
First, we leverage GPT-4o's strong multimodal reasoning and state-of-the-art vision specialists' precise perception to generate large-scale synthetic reasoning traces across diverse image sources.
Second, we generate reasoning traces using Python programs and structured reasoning templates, comparing them against GPT-generated traces to evaluate reasoning quality.  
In total, we produce over 1M reasoning traces across \dsnum datasets with GPT-4o and handcrafted programs. We then further apply filtering and mixing techniques and perform extensive experiments with different data ablations.

With the filtered 293K multi-modal reasoning traces, we finetune small 7-8B vision-language models to reason with vision specialists and evaluate our models on \benchmarknum benchmarks covering both perception and reasoning skills. We highlight four major takeaways from our experiments:
First, learning to reason with vision specialists enables our model to outperform vanilla instruction-tuned baseline by significant margins on both perception and reasoning benchmarks, with an overall average gain of 6.4\%. By contrast, the other distillation methods lead to smaller gains or even degradation in the perception performance. 
Second, our method consistently outperforms the vanilla instruction-tuned baseline by $4-5\%$ on average across all benchmarks regardless of model backbones, with staggering performance gains of $10-20\%$ on MMVet.
Third, through data ablations, we confirm that the quality of \cota\ matters more than quantity: our best data recipe consists of only 293K \cota\ which GPT-4o generated and answered correctly, and it leads to larger performance gains than other data recipes of larger scales. Finally, programmatically-generated \cota\ can hurt model performance as a result of the worse reasoning quality, suggesting that again that high-quality reasoning is crucial to the model's performance.

To summarize, we highlight three contributions: (1) We introduce a novel and the largest dataset of 293K multi-modal reasoning traces that cover 31 diverse data sources and include both single- and multi-image questions as well as image-text interleaved traces; (2) We demonstrate the effectiveness of our multi-modal reasoning data and showcase sizable performance gains over baselines on \benchmarknum\ benchmarks through extensive experiments; 
(3) Finally, our ablation studies reveal new insights into what matters in multi-modal reasoning data. We will release all artifacts publicly.

\begin{figure*}[ht]
    \centering
    \includegraphics[width=\textwidth]{images/framework.pdf}
    \vspace{-10pt}
    \caption{\textbf{Overview.} We propose \modelname: learning vision-language models to think with vision specialists via synthetic multi-modal reasoning traces.}
    \label{fig:framework}
    \vspace{-10pt}
\end{figure*}
\section{Related work}
We contextualize our work on multi-modal language models and multi-modal tool use.
\label{sec:related_work}

\noindent\textbf{Multi-modal language models.}
Recently, there have been many advances on open-source multi-modal models~\cite{awadalla2023openflamingo, chen2023internvl,  liu2023llava, liu2023improvedllava, liu2024llavanext, dai2024instructblip, li2022blip, li2023blip2, deitke2024molmo}. 
These efforts include training multi-modal models to take in multiple images, engage in multi-turn conversations, and even understand videos \cite{liu2024llavanext, jiang2024mantis, li2024llava-ov}. For example, LLaVA-Next achieves strong multi-image understanding through large-scale interleaved visual instruction tuning with M4-Instruct \cite{liu2024llavanext}. 
Similarly, Mantis introduces a new large-scale multi-image instruction tuning dataset Mantis-Instruct for multi-image training \cite{jiang2024mantis}. These efforts pave the foundation for our work on learning vision-language models with image-text interleaved reasoning traces.

\noindent\textbf{Multi-modal tool-use.}
Recently, there is growing interest in training multi-modal language models to be better at tool use~\cite{liu2023llavapluslearningusetools, qi2024cogcomtrainlargevisionlanguage, shao2024visualcot}. LLaVa-Plus first shows the possibility of training a multi-modal model to use vision specialists~\cite{liu2023llavapluslearningusetools}. Visual Program Distillation distills tool-use and reasoning abilities into a multi-modal model with chain-of-thought (CoT) data obtained from programs \cite{hu2024visual}. Similarly, Visual CoT introduces a new synthetic CoT dataset for training multi-modal models for enhanced reasoning \cite{shao2024visualcot}. More recently, LLaVa-CoT integrates both perception and reasoning from GPT-4o~\cite{xu2025llavacotletvisionlanguage}. Another closely related work CogCoM identifies 6 useful manipulations and trains multi-modal models with synthetic chain-of-manipulation (CoM) data \cite{qi2024cogcomtrainlargevisionlanguage}. Nonetheless, the manipulations are limited, and the authors only experiment with 70K CoM data.

Although these works demonstrate effectiveness, the proposed reasoning datasets are limited in scale and diversity, and none contains multi-image questions or includes images in the reasoning chains (Appendix~\ref{app:comparison} Table \ref{tab:datasets-comparison}). To complement existing works, we introduce a new large-scale dataset of 293K multi-modal interleaved reasoning traces that cover 31 data sources and include both single-image and multi-image questions.

\section{\modelname: Learning to Think with Vision Specialists}
\begin{figure*}[t]
    \centering
    \includegraphics[width=\textwidth]{images/data_gen.pdf}
    \vspace{-10pt}
    \caption{\textbf{Data generation.} We illustrate our model-based data generation (top) and programmatic generation (bottom) pipelines. 
    }
    \vspace{-10pt}
    \label{fig:dataset-gen}
\end{figure*}

Our goal is to train vision-language models to reason about complex multi-modal tasks with the help of vision specialists. To train such models, we need reasoning traces that involve (1) invoking vision specialists and (2) reasoning over their outputs. We refer to such data as \cota. 
We define a \cota\ $\mathcal{T}$ as a sequence of steps $S_i$, where each step consists of thought $t_i$, action $a_i$ and observation $o_i$: 
\vspace{0mm}
\begin{align}
  \mathcal{T} & = (S_0, S_1, ..., S_n) = (S_i)_{i=0}^{n} \\  
  S_i & = (t_i, a_i, o_i), t_i \in L, a_i \in A 
\end{align}

\vspace{0mm}
where $L$ represents language space, and $A$ is the action space consisting of vision specialists. The model only generates $t_i$ and $a_i$, which the training loss is applied on, whereas $o_i$ is obtained from the vision specialists.

\noindent\textbf{Action space.}
The action space $A$ of our model consists of vision tools that are either specialized vision models or image processing tools. Concretely, these include \tool{OCR} \cite{jadedai2025easyocr}, \tool{GetObjects} \cite{zhang2023recognize}, \tool{LocalizeObjects} \cite{liu2023grounding}, \tool{EstimateObjectDepth}, \tool{EstimateRegionDepth} \cite{depth_anything_v2}, \tool{DetectFaces} \cite{li2018dsfd}, \tool{Crop}, \tool{ZoomIn}, \tool{GetImageToTextsSimilarity}, \tool{GetImageToImagesSimilarity}, \tool{GetTextToImagesSimilarity}~\cite{Radford2021clip}. Inspired by prior works~\cite{hu2024visualsketchpadsketchingvisual, Gupta2022VisProg, ma2024mms}, we include a few additional tools to help with reasoning: \tool{QueryLanguageModel}, \tool{QueryKnowledgeBase}, \tool{Calculate}, and \tool{SolveMathEquation}. We also include \tool{Terminate} as a tool for the model to output a final answer in the same format. See the Appendix~\ref{app:actions} for all tools' implementation details. 

\subsection{\cota\ generation}
We generate synthetic \cota\ data with two automatic approaches: Model-based generation and Programmatic data generation.

\noindent\textbf{Model-based generation.}
This pipeline consists of three steps (Figure~\ref{fig:dataset-gen} top):

\datastep{1. Generate}. First, we leverage images and QA examples in existing visual instruction tuning datasets and generate \modelname-traces to solve the questions with GPT-4o (2024-08-06). We include diverse questions on both single-image and multi-image examples from two large-scale instruction tuning datasets, Cauldron and Mantis-Instruct~\cite{jiang2024mantis, laurençon2024idefics2}. We feed the images and questions to GPT-4o and prompt it to answer the questions by following a \cota\ or just CoT when it is not necessary (\eg,  the question is straightforward) or not helpful (\eg,  the question requires domain-specific knowledge) to call specialized vision tools (Figure \ref{fig:dataset-gen}). 
    We adopt ReAct-style prompting with JSON-format for calling the vision specialists and provide detailed instructions and examples in the prompt~\cite{yao2023reactsynergizingreasoningacting}. All prompts are in the Appendix~\ref{app:model_data_prompt}. 

 \datastep{2. Verify}. Second, we verify GPT-4o's generated answers against the ground-truth. We force GPT-4o to always end with \tool{Terminate}(answer) and compare its prediction to the ground-truth. If the final answer is correct, we move this \cota\ to the next stage. Otherwise, we convert this example into the direct answer (\da) format with the ground-truth (Figure~\ref{fig:dataset-gen}). 

\datastep{3. Parse}. Finally, we check the JSON syntax of each step of the \cota. Similar to the previous stage, we again keep the \cotas\ free of errors and turn the others into the \da\ format with ground-truth answers.


\noindent\textbf{Programmatic data generation.}
\begin{figure*}[ht]
    \centering
    \includegraphics[width=\textwidth]{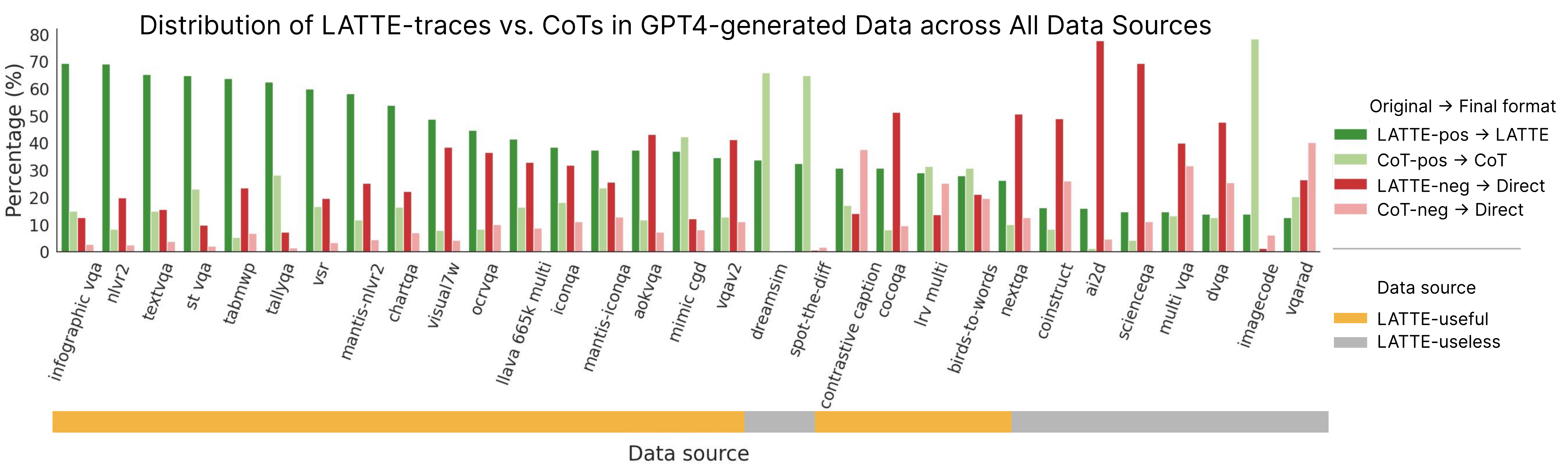}
    \vspace{-10pt}
    \caption{\textbf{Distribution of data formats and sources.} We visualize the frequency of data formats (i.e. \modelname-pos/neg, and \cotht-pos/neg, pos = correct final answers, neg = incorrect) in the original GPT-4-generated data and in our training data (i.e. \cota, \cotht, or \da) across all data sources. We also highlight the \modelname-useless (i.e. \% of \cotht-pos $-$ \modelname-pos $> 10$ or \% of \modelname-neg $-$ \modelname-pos $> 10$) vs. \modelname-useful datasets.}
    \label{fig:cota-da-dist}
    \vspace{-10pt}
\end{figure*}
In addition to distilling reasoning from proprietary models, we implement a programmatic data generation engine for synthesizing \cotas\ (Figure~\ref{fig:dataset-gen} bottom) and experiment with these data. 
This pipeline involves two steps:

 \datastep{1.Annotate}. First, we gather existing dense annotations of images. We adopt Visual Genome (VG) as it contains rich human annotations of objects, attributes, and relationships of the images. In addition, we obtain depth maps of the VG images with Depth-Anything-v2 \cite{depth_anything_v2}.
    
\datastep{2. Generate}. Next, we programmatically generate both the QA pairs and the corresponding \cotas\ with manually written templates and dense image annotations. We reuse the pipeline from \cite{zhang2024task,provision} to generate QA pairs that cover various vision capabilities such as counting and spatial understanding (See Appendix~\ref{app:thought_templates} for details). To generate \cotas, we define templates for thoughts, actions, and observations across all steps and fill in the templates with the collected annotations.  In particular, we manually design five thought templates for each action and randomly sample one during generation.  As for actions, we manually select the specialized vision tools for each type of questions (\eg, \tool{EstimateObjectDepth} for questions on objects' relative depths, and \tool{Localize} for object counting questions, etc.) and compose templates with them. 

\subsection{Data filtering and mixing}
We develop 3 filtering/mixing techniques, where we vary the distribution of: (1) data formats; (2) data sources; and (3) model- vs. program-generated reasoning traces. 

\noindent\textbf{Data format.} Model-generated data can be categorized into two formats: \cota\ or \cotht\ examples (Figure~\ref{fig:dataset-gen}). Additionally, they are further grouped into \cota/\cotht-pos and \cota/\cotht-neg examples where the final answers are correct and wrong respectively (Figure~\ref{fig:cota-da-dist}). Note that we convert both \cota-neg and \cotht-neg examples into the \da\ format with ground-truth answers (Figure~\ref{fig:dataset-gen}) so the final data format is one of \cota, \cotht, and \da. 

\noindent\textbf{Data source.} 
We also perform filtering based on data sources as Cauldron and Mantis-Instruct cover a wide range of tasks, some of which benefit more from vision specialists than others. 
To this end, we define \modelname-useless datasets as the ones where GPT-4o either decides to output \cotht\ much more often than \cota\ (i.e. \% of \cotht-pos $-$ \cota-pos $> 10$), or reaches wrong answers much more frequently than correct ones when using \cota\ (i.e. \% \cota-neg $-$ \cota-pos $> 10$) (Figure~\ref{fig:cota-da-dist}). The remaining datasets are considered \modelname-useful datasets.

\noindent\textbf{Program-generated data.} As the distribution of actions in model-generated data is imbalanced, with a couple of actions such as \tool{GetObjects} and \tool{OCR} dominating the dataset, we also try increasing action diversity by adding programmatic traces with underrepresented actions such as \tool{LocalizeObjects}, and \tool{EstimateRegionDepth}. 

\vspace{0mm}
\section{Experiments}
\begin{table*}[ht]
\centering
\caption{\textbf{\modelname\ vs. Vanilla IT with Different Models.} \modelname\ leads to performance gains over Vanilla IT regardless of the base models. The gains are 4-5\% on average across all \benchmarknum benchmarks and up to 17\% on MMVet.}
\label{tab:results_across_models}
\vspace{-5pt}
\resizebox{\textwidth}{!}{%
{\renewcommand{\arraystretch}{1.1}
\begin{tabular}{lll|cccc|cccc|l}
\toprule
\multirow{2}{*}{Language / Vision}  & \multirow{2}{*}{\makecell{Starting \\checkpoint}}      & \multirow{2}{*}{Method}   & \multicolumn{4}{l|}{\textbf{Perception}}        & \multicolumn{4}{l|}{\textbf{Perception + Reasoning}} & \textbf{Overall} \\ \cmidrule{4-12}
                                    &                                           &                           & CV-Bench & BLINK & RealWorldQA & Avg  & MathVista    & MMStar   & MMVet   & Avg    & Avg     \\ \midrule
\multirow{2}{*}{LLaMA3-8B / CLIP}   & \multirow{4}{*}{\makecell{Mantis\\Pretrained}}        & Vanilla IT                & 52.6     & 45.8  & 52.3        & 50.2 & 33.1         & 36.7     & 28.9    & 32.9   & 41.6    \\
                                    &                                           & \modelname & 56.9     & 49.6  & 51.1        & 52.6 & 36.6         & 40.8     & 45.2    & 40.8   & 46.7 ({\color{Green}+5.1})   \\ \cmidrule{1-1} \cmidrule{3-12}
\multirow{4}{*}{LLaMA3-8B / SigLIP} &        & Vanilla IT                & 52.3     & 43.7  & 51.8        & 49.3 & 31.1         & 40.5     & 33.0    & 34.9   & 42.1    \\
                                    &                                           & \modelname & \underline{57.2}     & 47.8  & 53.7        & 52.9 & 34.9         & 44.6     & 45.2    & 41.6   & 47.2 ({\color{Green}+5.1})    \\ \cmidrule{2-12}
                                    & \multirow{2}{*}{\makecell{Mantis\\Instruct-tuned}} & Vanilla IT                & 50.6     & 46.7  & 54.8        & 50.7 & 36.2         & 40.7     & 29.7    & 35.5   & 43.1    \\
                                    &                                           & \modelname & 51.7     & 47.3  & 56.1        & 51.7 & 38.9        & 45.1     & \underline{50.0}    & \underline{44.7}   & 48.2 ({\color{Green}+5.1})   \\ \midrule
\multirow{2}{*}{Qwen2-7B / SigLIP}  & \multirow{2}{*}{\makecell{LLaVa-OV \\Stage 1.5}}       & Vanilla IT                & 56.8     & \underline{50.3}  & \underline{57.8}        & \underline{55.0} & \underline{42.4}     & \underline{50.1}     & 39.3    & 43.9   & \underline{49.5}    \\
                                    &                                           & \modelname & \textbf{60.2}     & \textbf{52.6}  & \textbf{61.1}        & \textbf{58.0} & \textbf{46.9}        & \textbf{50.8}     & \textbf{50.9}    & \textbf{51.2}   & \textbf{53.8} ({\color{Green}+4.3})  \\ \bottomrule

\end{tabular}
}}
\end{table*}
\begin{table*}[ht]
\centering
\caption{\textbf{\modelname\ vs. Distillation Baselines.} \modelname\ brings substantial gains over the Vanilla IT baseline on both perception and perception + reasoning benchmarks, whereas VPD and LLaVa-CoT result in smaller gains. LLaVa-CoT even suffers from performance drop in perception tasks. All models were trained with 98K data.}
\label{tab:main_results}
\resizebox{\textwidth}{!}{%
{\renewcommand{\arraystretch}{1.1}
\begin{tabular}{l|cccl|cccl|l}
\toprule
\multirow{2}{*}{Method}    & \multicolumn{4}{l|}{\textbf{Perception}}        & \multicolumn{4}{l|}{\textbf{Perception + Reasoning}} & \textbf{Overall} \\ \cmidrule{2-10}
    &  BLINK & CV-Bench & RealWorldQA & Avg  & MathVista    & MMStar   & MMVet   & Avg    & Avg     \\ \midrule
Vanilla IT & \underline{44.1}  & \underline{49.2}     & 41.4        & 44.9 & 31.0         & 39.7     & 27.8    & 32.8   & 38.9    \\
VPD        & 41.6  & 48.8     & \textbf{44.8}        & \underline{45.1} ({\color{Green}+0.2}) & 33.0         & 41.1     & 32.8    & 35.7 ({\color{Green}+2.8})  & \underline{40.4}  ({\color{Green}+1.5})  \\
LLaVa-CoT  &  42.2  & 40.4     & 38.0        & 40.2 ({\color{BrickRed}-4.7})& \underline{36.7}         & \textbf{44.6}     & \underline{40.2}    & \underline{40.5} ({\color{Green}+7.7})  & \underline{40.4} ({\color{Green}+1.5})  \\
\modelname       & \textbf{46.4}  & \textbf{54.0}     & \underline{42.0}        & \textbf{47.5} ({\color{Green}+2.6})& \textbf{36.9}         & \underline{44.2}     & \textbf{47.9}    & \textbf{43.0} ({\color{Green}+10.2})  & \textbf{45.2}  ({\color{Green}+6.4})\\ \bottomrule 

\end{tabular}
}}
\vspace{-5pt}
\end{table*}

We perform extensive experiments with small 7-8B multi-modal models and various data recipes on \benchmarknum benchmarks to study two questions: (1) do \cotas\ improve small vision-language models' performance on both perception and reasoning VQAs? (2) what matters in \cotas? 

\noindent\textbf{Models.} 
We adopt models with multi-image support as our reasoning traces include multiple images. For most experiments, we use Mantis-8B-SigLIP-LLaMA-3 as the base model. We additionally experiment with Mantis-8B-CLIP-LLaMA-3, and LLaVA-OneVision-7B (Qwen2-7B and SigLIP) to showcase our method's generalizability. 

\noindent\textbf{Baselines.}
We compare \modelname\ to three types of baselines: (1) vanilla instruction-tuning (IT): instruction-tuning with only direct answers; (2) distillation methods that distill both perception and reasoning from larger models into smaller models, including VPD \cite{hu2024visual}\footnote{As VPD is close-sourced, we reproduce their data by converting \cotas\ into CoTs in VPD's format.}, LLaVa-CoT~\cite{xu2025llavacotletvisionlanguage}, and VisCoT \cite{shao2024visualcot}\footnote{Since VisCoT only has reasoning steps for one data source GQA, training with its data leads to much worse performance. We include its results in the Appendix~\ref{app:vis_cot_performance}.}; 
For fair comparison, we train our models and baselines with the same base model, the same hyperparameters, and the same number of examples; (3) multi-modal agents that use tools at inference time, including LLaVa-Plus~\cite{liu2023llavapluslearningusetools} and CogCoM~\cite{qi2024cogcomtrainlargevisionlanguage}.

\noindent\textbf{Training details.} 
We finetune models starting from checkpoints at different stages -- pretrained and instruction tuned for Mantis-8B-SigLIP-LLaMA-3, and stage 1.5 for LLaVA-OneVision-7B -- to investigate if and where \cotas\ bring gains. We adopt the hyperparameters from \cite{liu2024llavanext, jiang2024mantis} and fine-tune both the language model and the projector with learning rate $=1e-5$ for 1 epoch with either NVIDIA A100s 40GB or H100s 80GB. We additionally perform hyperparameter tuning with LLaVA-OneVision-7B and include this result in the Appendix~\ref{app:add_train_details}. 

\begin{figure*}[t]
    \centering
    \includegraphics[width=0.9\textwidth]{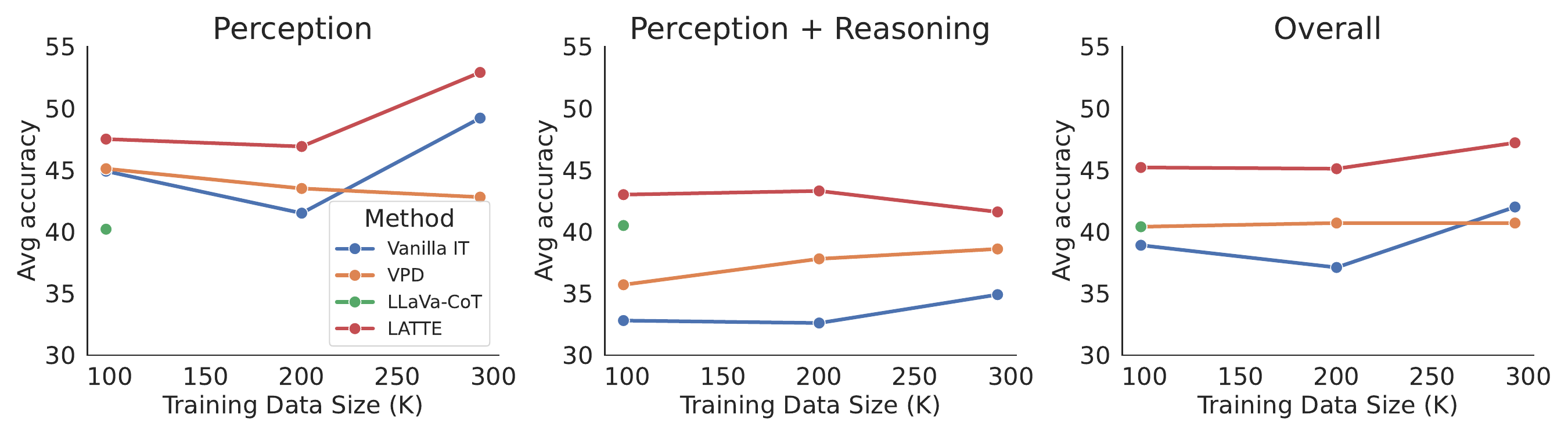}
    \caption{\textbf{\modelname\ vs. Distillation Baselines across Training Data Scales.} \modelname\ leads to consistent gains on perception and reasoning benchmarks over the Vanilla instruction-tuned baseline across varying training data sizes -- 98K, 200K and 293K -- and the gains are larger than VPD's. LLaVa-CoT only has 98K data.}
\vspace{-1mm}
    \label{fig:scale-results}
\end{figure*}
\begin{figure*}[ht]
    \centering
    \includegraphics[width=\textwidth]{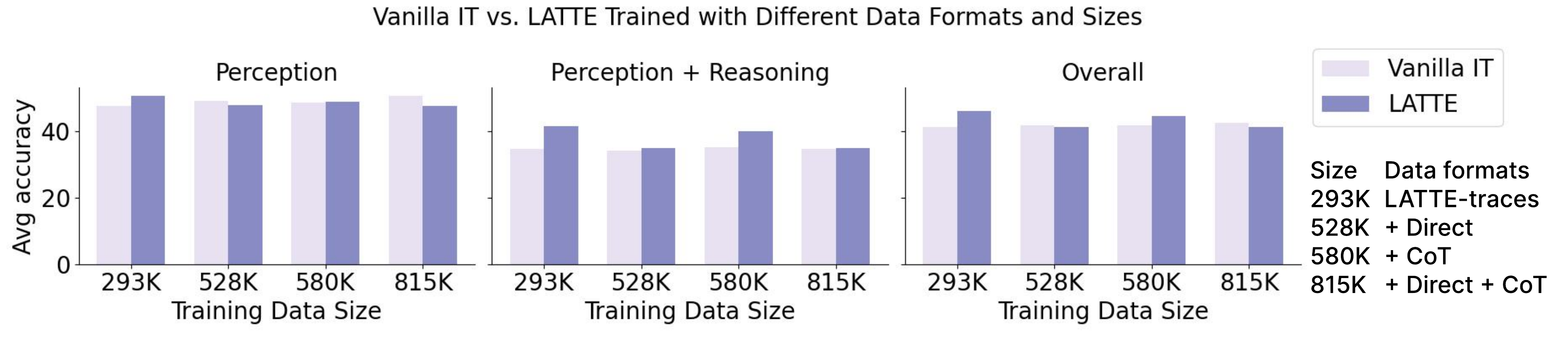}
    \caption{\textbf{Ablations on Data Formats.} 293K \cotas\ lead to the greatest gains over Vanilla IT and the highest overall performance. Adding either \cotht~or \da\ doesn't bring additional gains despite the increased size. }
    \label{fig:ablations-across-formats}
    \vspace{-5pt}
\end{figure*}
\noindent\textbf{Evaluation setup.}
We select \benchmarknum VQA benchmarks covering both perception and reasoning. The perception-focused benchmarks include RealWorldQA, CV-Bench and BLINK \cite{tong2024eyes, schwenk2022aokvqabenchmarkvisualquestion, fu2024blink, li2023seed2}. 
We also include 3 benchmarks that additionally test reasoning capabilities: MathVista, MMStar, and MMVet \cite{lu2024mathvista, chen2024mmstar, yu2024mmvet}. We adapt VLMEvalKit \cite{duan2024vlmevalkit} for our evaluation, where an LLM judge (i.e. GPT-4-turbo) is used to score predictions between 0 and 1 compared to the groundtruth short answers for open-ended questions. Additional details are in Appendix~\ref{app:add_eval_details}. 

\subsection{Do \cotas\ improve models' performance on both perception and reasoning VQAs?}


\paragraph{\modelname\ beats Vanilla IT on average across all benchmarks regardless of the base model and checkpoint, with significant gains of up to 17\% on MMVet.}
We fine-tune 3 different multi-modal models with all 293K \cotas\ starting from different checkpoints. We observe that our method leads to consistent gains of 4-5\% in the model's average accuracy across \benchmarknum benchmarks compared to the baselines instruction-tuned with the same examples in the \da\ format (Table~\ref{tab:results_across_models}). We note that our method results in staggering gains of up to 17\% on MMVet, which covers a wide range of perceptual and reasoning capabilities. 

\noindent\textbf{\modelname\ leads to substantial gains over vanilla instruction-tuning on both perception and reasoning benchmarks, whereas distillation baselines result in smaller gains or even degradation on some perception tasks.} 
We find that learning to reason with vision specialists enables our model to achieve consistent gains on perception-focused VQA benchmarks as well as benchmarks that require both perception and reasoning, with average gains of 2.6\% and 10.2\% respectively (Table \ref{tab:main_results}). By contrast, both distillation baselines VPD and LLaVa-CoT bring much smaller gains, with an average of 1.5\% across all benchmarks, compared to ours (6.4\%). Further, we observe that the same trend holds as we scale the training data size from 98K to 200K and 293K, where our method consistently brings larger gains on both perception and perception + reasoning benchmarks (Figure \ref{fig:scale-results}). Interestingly, LLaVa-CoT even hurts the model's performance on perception benchmarks, even though it increases the performance on the perception + reasoning benchmarks (Table \ref{tab:main_results}). This result suggests that GPT4-o might still be inferior to vision specialists on some perception tasks, as LLaVa-CoT distills purely from GPT4-o.

\vspace{0mm}
\begin{table}[ht]
\centering
\caption{\textbf{\modelname\ vs. Multi-modal Agent Baselines.} Training with \cotas\ leads to much larger gains.}
\label{tab:agent_baselines}
\vspace{0mm}
\resizebox{0.5\textwidth}{!}{%
{\renewcommand{\arraystretch}{1.1}
\begin{tabular}{lc|ll}
\toprule
\multirow{2}{*}{Model}                  & \multirow{2}{*}{Data size}  & \multicolumn{2}{l}{Base model $\rightarrow$ Finetuned model} \\ \cmidrule{3-4}
                  & & MathVista & MMVet \\ \hline
LLaVA-Plus                & 158K                              & ---       & 32.5 $\rightarrow$ 35.0 (+2.5)   \\
CogCoM                    & 70K                              & 34.8 $\rightarrow$ 35.7 (+0.9)    & 45.9 $\rightarrow$ 46.1 (+0.2) \\

\modelname & 98K                              & 32.7 $\rightarrow$ 36.9 (\underline{+4.2})   & 34.4 $\rightarrow$ 47.9 
 (\underline{+13.5})\\ 
 \modelname & 293K                               & 32.7 $\rightarrow$ 38.9 (\textbf{+6.2})   & 34.4 $\rightarrow$ 50.0 (\textbf{+15.6}) \\
 \bottomrule
\end{tabular}
}}
\vspace{-10pt}
\end{table}
\noindent\textbf{\modelname\ scores higher on MathVista and MMVet than multi-modal agent baselines do, and \cotas\ bring larger gains to the base model.} We see in Table \ref{tab:agent_baselines} that \modelname\ achieves higher accuracies on MathVista and MMVet. Moreover, \cotas\ bring much larger gains to the base model than LLaVa-Plus and CogCoM's data do, despite its comparable size.   

\vspace{0mm}
\subsection{What matters in \cotas?}
We perform ablations with \cotas\ to study what matters in improving models' performance. 
For model-generated data, we explore two data filtering techniques on (1) data formats and (2) data sources
(Figure~\ref{fig:cota-da-dist}).
\vspace{0mm}

\begin{table}[ht]
\centering
\caption{\textbf{Ablations on data sources.} Including all sources hurts model's perception and overall performance while having only \modelname-useful datasets helps.}
\label{tab:ablations-arcoss-data-sources}
\vspace{0mm}
\resizebox{0.48\textwidth}{!}{%
{\renewcommand{\arraystretch}{1.1}
\begin{tabular}{llllll}
\toprule
Data source                                                & Size                  & Method                    & \textbf{Percept.}   & \textbf{P. + Reason.} & \textbf{Overall}     \\ \midrule
\multirow{2}{*}{All datasets}                              & \multirow{2}{*}{815K} & Vanilla IT                & 50.7        & 34.7                   & 42.7        \\
                                                           &                       & \modelname & 47.7 ({\color{BrickRed}-3.0}) & 35.1 ({\color{Green}+0.4})            & 41.4 ({\color{BrickRed}-1.4}) \\ \midrule
\multirow{2}{*}{\modelname-useful datasets} & \multirow{2}{*}{566K} & Vanilla IT                & 46.3        & 33.3                   & 39.8        \\
                                                           &                       & \modelname & 46.8 ({\color{Green}+0.5}) & 35.6 ({\color{Green}+2.3})            & 41.2 ({\color{Green}+1.4}) \\ \bottomrule
\end{tabular}
}}
\vspace{-10pt}
\end{table}
\noindent\textbf{Data quality matters more than quantity: 293K \cotas\ lead to higher performance than larger mixtures of \cotas\ and \cotht\ or \da.} 
 We find that 293K \cotas\ result in the biggest gain of 5\% on average over the baseline across all benchmarks (Figure ~\ref{fig:ablations-across-formats}). Adding \cotht\ examples results in a smaller gain of 2.6\%, even though the training data size almost doubles (Figure~\ref{fig:ablations-across-formats}). On the other hand, combining \cota\ and \da\ examples hurts the model's performance compared to \cotas\ only, especially on the perception tasks (Figure \ref{fig:ablations-across-formats}). 
 We empirically observe that models trained with a mix of \cotas\ and \da\ examples tend to adopt the \da\ format more often (around 70\%) at inference time, relying on its own weaker perceptual ability instead of vision specialists' and thus scoring lower. 

\noindent\textbf{Data sources matter too: including all datasets hurts performance while including only \modelname-useful datasets brings gains.}
Similarly, we see that including only the \modelname-useful datasets -- where GPT-4o frequently chooses to use vision specialists and reaches correct final answers -- improves the model's average performance compared to the baseline, while including all data sources doesn't (Table~\ref{tab:ablations-arcoss-data-sources}). Again, we see that a smaller set of 566K \cotas\ leads to better performance than a much larger dataset (815K), implying that data quality matters more than quantity.

\begin{table}[ht]
\centering
\caption{\textbf{Ablations on programmatic \cotas.} We find that training with additional programmatic \cotas\ doesn't bring more gains.}
\label{tab:program-data-ablations}
\vspace{-5pt}
\resizebox{0.48\textwidth}{!}{%
{\renewcommand{\arraystretch}{1.1}
\begin{tabular}{lll|ccc}
\toprule
M: P & Data format & Size & \textbf{MathVista} & \textbf{Percept. + Reason.} & \textbf{Overall} \\ \midrule
---    & Direct      & \multirow{3}{*}{293K} & 31.1       & 34.9                   & 42.0    \\
0:1    & P-traces      &                       & 17.3       & 15.9                   & 27.2    \\
1:0    & M-traces     &                       & 34.9     & \underline{41.6}                 & \textbf{47.2}    \\ \midrule
1:0.1  & +P-traces 29K  & 322K                  & 33.9      & 40.1                   & 44.0    \\
1:0.25 & +P-traces 73K  & 366K                  & \textbf{38.3}       & \textbf{42.1}                   & \underline{46.3}    \\
1:0.5  & +P-traces 147K & 440K                  & \underline{36.7}       & 39.7                   & 45.5    \\
1:1    & +P-traces 293K & 586K                  & 31.0       & 36.2                   & 43.2   
\\ \bottomrule    

\end{tabular}
}}
\vspace{-5pt}
\end{table}
\noindent\textbf{Programmatically generated \cotas\ can help on a certain benchmark but not overall, likely due to the worse quality of thoughts.}
We experiment with a mixture of model-generated and programmatic reasoning traces, with ratios ranging from 1:0.1 to 1:1. We find that training with only programmatic \cotas\ results in large performance drops (Table~\ref{tab:program-data-ablations}). Similarly, while adding programmatic \cotas\ can bring gains on some benchmark (e.g. MathVista), it fails to bring overall gains despite the increased data size (Table~\ref{tab:program-data-ablations}). This is likely due to the model's worse reasoning capability learned from templated thoughts. See more details in Appendix~\ref{app:error_analysis} (Figure~\ref{fig:mcota-and-pcota-examples}).

Overall, our experiments suggest that the quality of perceptual information and reasoning are both crucial to improving vision-language models' performance across diverse VQAs. 

\subsection{Additional ablations}
\begin{figure*}[h]
 \vspace{-3mm}
    \centering
    \includegraphics[width=\textwidth]{images/error_analysis.pdf}
    \vspace{-5mm}
    \caption{\textbf{Qualitative analysis.} Example outputs of VPD, LLaVA-CoT vs. \modelname\ on BLINK.}
    \label{fig:latte_vs_distillation_error_analysis}
    \vspace{-3mm}
\end{figure*}
\begin{table}[ht]
\vspace{-5pt}
\centering
\caption{\textbf{Ablations on \modelname's inference setup.} The OCR tool greatly affects model's performance, while the query LLM tool doesn't; and increasing the maximum number of tool calls doesn't help beyond 10.}
\label{tab:inference-ablations}
\vspace{-5pt}
\resizebox{0.48\textwidth}{!}{%
{\renewcommand{\arraystretch}{1.1}
\begin{tabular}{llll}
\toprule
Method                    & \textbf{Percept.} & \textbf{P. + Reason.} & \textbf{Overall} \\ \hline
LATTE (max 10 calls) & 51.7       & 43.8      & \textbf{47.8}    \\

max 5 calls           & 51.7       & 42.8      & 47.2    \\
max 20 calls            & 51.6       & 43.6      & 47.6    \\ \hline
no \tool{QueryLM}         & 52.1       & 43.5      & \textbf{47.8}    \\
\tool{OCR} with easyocr          & 51.4       & 39.9      & 45.7   \\ \bottomrule
\end{tabular}
}}
\vspace{-5pt}
\end{table}
\noindent\textbf{What matters in \modelname's inference setup?}
In addition to training data, we also perform ablations on  \modelname's inference setups, including tool ablations and tool call number ablations. 

First, the tool ablations include (1) excluding the \tool{QueryLanguageModel} tool, and (2) using EasyOCR ~\cite{jadedai2025easyocr} instead of GPT for \tool{OCR}. We see that removing \tool{QueryLanguageModel} leads to no difference in the model's overall performance (Table~\ref{tab:inference-ablations}), suggesting that \tool{QueryLanguageModel} is not an important tool to the VQA benchmarks in our evaluation. Nevertheless, by using EasyOCR for \tool{OCR} we do see a large drop in the model's performance, especially on reasoning-intensive benchmarks such as MMVet due to EasyOCR's less accurate outputs (Table~\ref{tab:inference-ablations}). Still, \modelname\ is over 2\% above the vanilla instruction-tuned baseline on average across all benchmarks.  

Second, as we adopt multi-turn conversations for all models with our method, we can vary the number of tool calls at inference time. We set the maximum number of tool calls to 10 by default and experiment with 5 and 20 tool calls. Through this experiment, we see that the model's scores peak at 10 maximum tool calls, suggesting that 10 tool calls are sufficient in most cases (Table~\ref{tab:inference-ablations}). 

\begin{table}[ht]
\centering
\caption{\textbf{Ablations on \modelname's model backbone.} \cotas\ improve stronger base models too. }
\label{tab:model-backbone-ablations}
\vspace{-5pt}
\resizebox{0.48\textwidth}{!}{%
{\renewcommand{\arraystretch}{1.1}
\begin{tabular}{lllll}
\toprule
Model backbone                & Method     & \textbf{Percept.} & \textbf{P. + Reason.} & \textbf{Overall} \\ \hline
\multirow{2}{*}{InternVL3-8B} & Vanilla IT & 51.5       & 45.1      & 46.7    \\
                              & LATTE      & 54.1       & 54.5      & \textbf{52.0}    \\ \hline
\multirow{2}{*}{Qwen2.5VL-3B} & Vanilla IT & 54.3       & 49.6      & 50.8    \\
                              & LATTE      & 53.4       & 55.7      & \textbf{55.1}    \\ \hline
\multirow{2}{*}{Qwen2.5VL-7B} & Vanilla IT & 53.7       & 52.5      & 52.8    \\
                              & LATTE      & 56.9       & 57.7      & \textbf{57.5}   \\ \bottomrule
\end{tabular}
}}
\vspace{-10pt}
\end{table}

\noindent\textbf{Do \cotas\ improve stronger base models?}
We further ablate on the model backbone, as new and stronger open-source vision-language models come out.  Our results using the latest vision-language models -- Qwen2.5VL and InternVL3 -- as the base models demonstrate that our method improves upon vanilla instruction tuning even with strong base models (Table~\ref{tab:model-backbone-ablations}).

\vspace{0mm}
\subsection{Error Analysis}
\noindent\textbf{Where does \modelname\ perform better than the distillation baselines?}
We find that \modelname\ performs better in fine-grained perception tasks such as the counting questions in BLINK, while VPD and LLaVA-CoT tend to hallucinate and make perceptual errors (Figure~\ref{fig:latte_vs_distillation_error_analysis}).

\begin{table}[ht]
\vspace{-5pt}
\centering
\caption{\textbf{\modelname's Error Types on MMVet.}}
\label{tab:mmvet_error_analysis}
\vspace{-5pt}
\resizebox{0.35\textwidth}{!}{%
{\renewcommand{\arraystretch}{1.1}
\begin{tabular}{lll}
\toprule 
Error type   &       Subtype         & \% \\ \hline
\multirow{2}{*}{Tool call}        & Format            & 3              \\
                                  & Value             & 6              \\ \hline
Tool result                       & ---               & 7              \\ \hline
\multirow{2}{*}{Model perception} & Not using tools   & 6              \\
                                  & Irrelevant tools  & 50             \\ \hline
Model reasoning                   & ---               & 28             \\ 
\bottomrule           
\end{tabular}
}}
\vspace{-5pt}
\end{table}
\noindent\textbf{What errors does \modelname\ make?}
On MMVet, we find that the model's most frequent error happens when it falls back on its own perceptual ability after deciding not to use tools or finding the vision tools' outputs irrelevant/not helpful for the question (\eg movie, arts, or medical questions that require domain knowledge) (Table~\ref{tab:mmvet_error_analysis}). 
These numbers suggest that the model's performance can be improved by diversifying tools and questions in the training data, and strengthening reasoning.



\section{Conclusion}

We propose to learn vision-language models to reason with vision specialists.
To learn such models, we synthesize a novel large-scale dataset of multi-modal reasoning traces grounded on perceptual information. With this data, we fine-tune small vision-language models and perform extensive experiments.  Across \benchmarknum benchmarks covering both perception and reasoning, we demonstrate that our model achieves significant gains over vanilla instruction-tuned baselines and other distillation methods in perception and reasoning tasks. 

\section{Limitations}
 First, our method requires customized implementations of the specialized vision tools. Second, reasoning with the vision specialists requires additional compute at inference time. Nevertheless, it is becoming a common practice to increase model performance by scaling up test time compute \cite{openai2025o3, muennighoff2025s1simpletesttimescaling}. Future work can optimize and enhance the implementations of vision specialists, especially as the computer vision community continues to advance vision models. Additionally, while we try to include the most important tools for general perception and reasoning, other types of VQA \eg knowledge-intensive ones might benefit from additional tools as suggested in our error analysis. Lastly, due to the limited generalization of supervised finetuning and diversity of the visual world, researchers might need to explore training alteratives (\eg reinforcement learning) for better generalization or train new models with different vision specialists for other applications (\eg web navigation) or for other domains (\eg medical question answering).  


\section{Acknowledgement}
Zixian Ma was partially funded by Sony for this project. Zixian Ma conducted most of this work at a Salesforce internship.

\bibliography{main}

\appendix
\clearpage
\setcounter{page}{1}

\section{Dataset and model comparison}

\begin{table*}[ht]
\vspace{-3mm}
\Large
\caption{\textbf{Dataset and model comparison.}}
\vspace{-3mm}
\label{tab:datasets-comparison}
\resizebox{\linewidth}{!}{%
{\renewcommand{\arraystretch}{0.9}
\centering
\begin{tabular}{p{4cm}|p{3.5cm}p{1.7cm}p{4cm}p{2.2cm}p{2.2cm}|p{2.5cm}p{2cm}}
\toprule
\multirow{3}{*}{Paper} & \multicolumn{5}{c|}{Dataset}                                                                                                                                                                                                                                                                                                                                                                                                                                                                                                                   & \multicolumn{2}{c}{Model}                             \\ \cmidrule{2-8}
                       & Training set size                                                              & Data source number                                                                                                             &  Tool number                                                                                                                                                                                                                       & Multi-image questions & Multi-modal reasoning chain* & Inference-time tool-use & Multi-image support \\ \midrule
Science QA~\cite{saikh2022scienceqa}             & 12.6K                                                                                  & 1                                                                                                              & \xmark                                                                                                                                                                                                                                                    & \xmark                     & \xmark                          & \xmark                             & \xmark                   \\ \midrule
M3CoT~\cite{chen2024m}                & 7.8K                                           & 2                                                                                                             & \xmark                                                                                                                                                                                                                                                    & \xmark                    & \xmark                         & \xmark                            & \xmark                  \\ \midrule

VPD~\cite{hu2024visual}                    & 90K                                           & 6                                                                              & 6                                                                                                                                                                                   & \xmark                    & \xmark                         & \xmark                            & \xmark                  \\ \midrule

LLaVA-CoT~\cite{xu2025llavacotletvisionlanguage}               & 100K                                                      & 10                                    & \xmark                                                                                                                                                                                                                                                    & \xmark                    & \xmark                         & \xmark                            & \xmark                  \\ \midrule
V*~\cite{wu2024v}                   & 206K                                           & 3                                                                              & 1                                                                                                                                                                                  & \xmark                    & \xmark                         & \xmark                            & \cmark                  \\ \midrule
LLaVA-Plus~\cite{liu2023llavapluslearningusetools}                   & 158K                                           & 6                                                                              & 12  (or 19 counting compositional ones)                                                                                                                                                                                  & \xmark                    & \xmark                         & \cmark                            & \xmark                  \\ \midrule
VisualCoT~\cite{shao2024visualcot}& 98K (+340K with bboxes but no thoughts)  & 12  & 2                                                                                                                                                                                                                                          & \xmark                    & \xmark                         & \cmark                            & \cmark                  \\\midrule

CogCoM~\cite{qi2024cogcomtrainlargevisionlanguage}               & 70K                                                  & 3                                    & 6                                                                                                                                                                                                                                                    & \xmark                    & \xmark                         & \cmark                            & \cmark                  \\ \midrule
\modelname                  & 293K (+over 1M program generated reasoning traces)                         & 31                                                                                                         & 15 tools (see Section 2) & \cmark                    & \cmark                         & \cmark                            & \cmark                \\ \bottomrule 
\multicolumn{8}{l}{*The reasoning chain contains not just texts but also images.} 
\end{tabular}
}
}
\end{table*}
\label{app:comparison}
We summarize the differences between our work and the other multi-modal CoT datasets including ScienceQA, M$^3$COT, Visual CoT, VPD, V*, LLaVa-Plus, LLaVA-CoT, CogCoM in Table~\ref{tab:datasets-comparison}.

\section{Additional results}
\subsection{Additional qualitative examples}
We present additional successful outputs of \modelname\ across both single-image and multi-image examples in Figures~\ref{fig:add-examples} and \ref{fig:add-examples-2} as well as failure cases in Figure~\ref{fig:failure-examples}.

\subsection{Qualitative error analysis}
\begin{figure*}[h]
    \centering
    \includegraphics[width=\textwidth]{images/mcota_and_pcota_examples.pdf}
    \vspace{-0pt}
    \caption{\textbf{Qualitative analysis.} Examples of \modelname\ success and failure after adding programmatic \cotas.}
    \label{fig:mcota-and-pcota-examples}
    \vspace{-0pt}
\end{figure*}
\label{app:error_analysis}
\noindent\textbf{Why does adding programmatic \cota\ help on MathVista but hurt MMVet performance?}
We observe that adding programmatic \cota\ can result in up to 3\% gain on MathVista and 9\% drop on MMVet. Upon analysis, we discover that programmatic \cota\ improves the general VQA split in MathVista the most by almost 9\%. This is because \tool{Localize} is helpful for these questions, and our programmatic data includes many \tool{Localize} instances that allow \modelname\ to learn to use it effectively (Figure~\ref{fig:mcota-and-pcota-examples}). Conversely, programmatic data hurts \modelname's performance on MMVet most likely due to the model's worse reasoning ability as a result of the simple and rigid thoughts generated with templates in our programmatic data (Figure~\ref{fig:mcota-and-pcota-examples}). 

\subsection{Additional quantitative results}
We report additional quantitative results of data ablations on Mantis-CLIP in Table~\ref{tab:model-data-ablations-clip}, where we see the same trends we observe with Mantis-SigLIP: the smallest dataset of 293K \cota\ examples leads to the highest absolute performance and gain compared to other datasets with a mix of \cota, \cotht, and/or \da\ examples at larger scales.

\begin{table}[h]
\Large
\caption{\textbf{VisCoT results}}
\label{tab:vpd-viscot-resutls}
\resizebox{\linewidth}{!}{%
\begin{tabular}{llllllllll}
Method                     & A-OKVQA & BLINK & MathVista & MMMU & MMStar & MMVet & MMVP & RealWorldQA & Avg  \\ \midrule
 VPD-LLaVA   & 80.7    & 43.9  & 35.1      & 40.8 & 40.1   & 40.6  & 61.3 & 45.1        & 48.5 \\
 VisCoT-LLaVA & 67.9    & 39.4  & 12.9      & 36.1 & 36.0   & 26.7  & 61.3 & 34.0        & 39.3
\end{tabular}
}
\end{table}

\noindent\textbf{Visual-CoT Performance.} 
\label{app:vis_cot_performance}
We experimentally compare \modelname\ to Visual-CoT.
We finetune Mantis-LLaVA-Pretrained (LLama3+SigLIP) with Visual CoT and compare its performance with \modelname\ (Table~\ref{tab:vpd-viscot-resutls}). We use 413K examples where the bounding boxes are valid and within the image. We find that the models trained with Visual CoT data achieve an average accuracy of 39.3\% (much lower as Visual COT's data are mostly Text/Doc images and contain only bboxes without natural language thoughts) on the benchmarks. 

\begin{table}[h]
\Large
\caption{\textbf{\modelname\ vs. Direct inference}}
\label{tab:cota-vs-direct}
\resizebox{\linewidth}{!}{%
\begin{tabular}{llllllllll}
Test prompt                     & A-OKVQA & BLINK & MathVista & MMMU & MMStar & MMVet & MMVP & RealWorldQA & Avg  \\ \midrule
\modelname  & 82.1 & 45.9 & 37.0 & 39.8 & 43.4 & 46.7 & 66.0 & 41.6 & 50.3 \\
Direct & 81.9 & 46.6 & 31.2 & 42.1 & 39.1 & 35.0 & 68.3 & 40.2 & 48.0
\end{tabular}
}
\end{table}
\noindent\textbf{Performance gain with \modelname\ inference.} 
We compare the model's performance when trained with a random mix of 293K \cotas\ and Direct data (1:1) and tested with \modelname\ format vs. Direct prompt. We find that the model achieves an average of 50.3\% when tested following \modelname\ format vs. 48\% with the Direct prompt (Table~\ref{tab:cota-vs-direct}), suggesting that reasoning with vision specialists at inference time improves model's performance.

\noindent\textbf{Hyperparameter tuning}
\begin{figure}[tb]
    \centering
    \includegraphics[width=0.9\columnwidth]{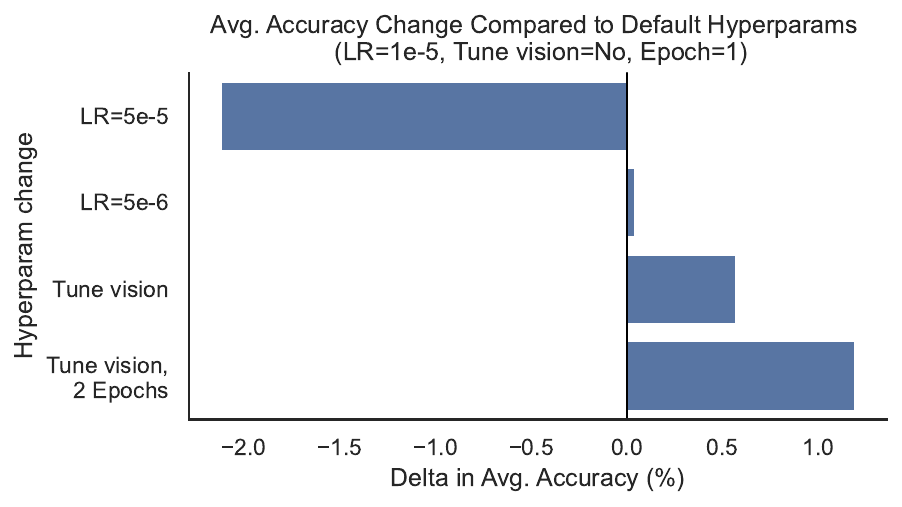}
    \vspace{-10pt}
    \caption{\textbf{Hyperparameter ablations.} Additional gains can be achieved with a smaller learning rate for the language model, tuning the vision encoder, and training for more epochs. }
    \label{fig:hyperparam-result}
    \vspace{-15pt}
\end{figure}
\noindent\textbf{Additional gains can be achieved by tuning the vision encoder, training with a smaller learning rate or for more epochs.}
Last but not least, our hyperparameter tuning experiments with LLaVa-OV-Stage1.5 suggest that we can further improve the model's absolute performance by tuning the vision encoder, training with a smaller learning rate and/or for longer epochs (Figure~\ref{fig:hyperparam-result}).

\section{Model-based data generation}
\label{app:model_data_prompt}
\subsection{Generation prompt}
We present the full data generation prompt used in our model-based data generation pipeline in Listing~\ref{lst:data-gen-prompt}.

\subsection{Dataset statistics}
We present a table with detailed statistics of the \cota\ 293K dataset in Table~\ref{tab:dataset-stats}. 

\section{Action implementation}
\label{app:actions}
Our Python implementation of all actions can be found in Listing~\ref{lst:actions}.

\section{Programmatic data generation}
\subsection{QA and action templates}
We present the question-answer and corresponding action templates used in our programatic data generation in
Table~\ref{tab:program-data-templates}. We design 16 different question templates for both single-image and multi-image examples that cover 5 capabilities: attribute recognition, counting, 2D and 3D spatial understanding, and multi-image understanding.

\subsection{Thought templates}
\label{app:thought_templates}
We also present the five thought templates in Listing~\ref{lst:thought-templates} we define for each action, where one of them is randomly sampled and used during generation. 

\subsection{Example action distribution}

\begin{figure*}[t]
    \centering
    \includegraphics[width=0.6\textwidth]{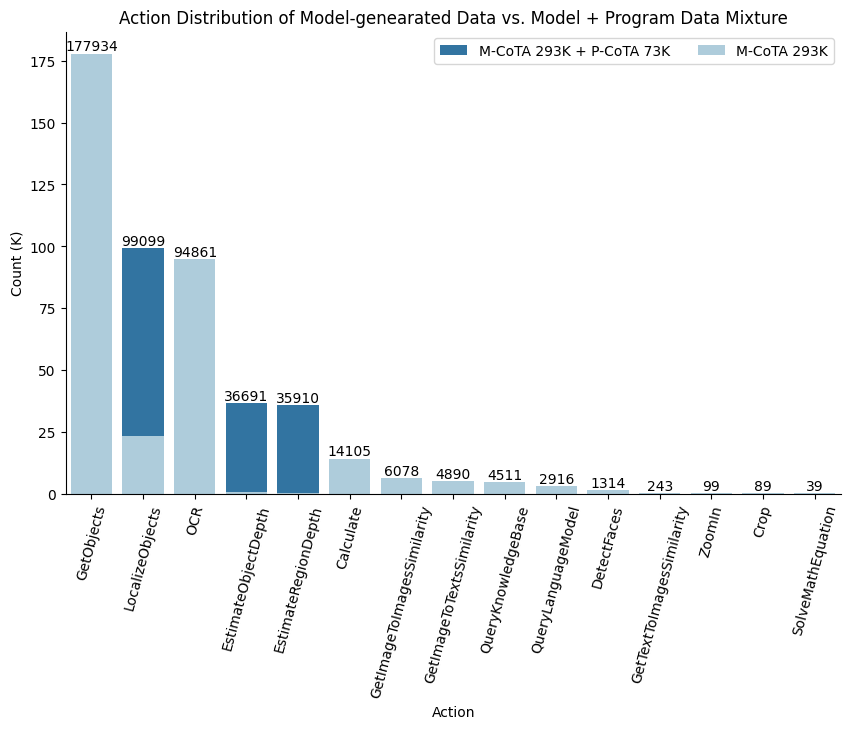}
    \caption{Action distribution of model-generated data vs. model and program data mixtures.}
    \label{fig:act-dist-before-and-after}
\end{figure*}

We plot example distributions of all actions before and after adding programmatic \cota\ 73K data in 
Figure~\ref{fig:act-dist-before-and-after}.

\section{Additional training details}
\label{app:add_train_details}
We report additional training hyperparameters for Mantis models and LLaVA-OV in Table~\ref{tab:add-train-hyperparams-mantis} and \ref{tab:add-train-hyperparams-llava-ov} respectively.

\begin{table}[H]
\centering
\caption{\textbf{Additional training hyperparameters for Mantis-SigLIP and Mantis-CLIP.}}
\label{tab:add-train-hyperparams-mantis}
\resizebox{0.8\columnwidth}{!}{%
{\renewcommand{\arraystretch}{1.0}
\begin{tabular}{ll}
Name                            & Value  \\ \midrule\midrule
bf16                            & TRUE   \\
tf32                            & True   \\
per\_device\_train\_batch\_size & 1      \\
per\_device\_eval\_batch\_size  & 1      \\
gradient\_accumulation\_steps   & 16     \\
weight\_decay                   & 0      \\
warmup\_ratio                   & 0.03   \\
lr\_scheduler\_type             & cosine \\
lora\_enabled                   & FALSE  \\
qlora\_enabled                  & FALSE  \\
max\_seq\_len                   & 8192  \\ \bottomrule
\end{tabular}
}
}
\vspace{-15pt}
\end{table}

\begin{table}[H]
\centering
\caption{\textbf{Additional training hyperparameters for LLaVA-OV.}}
\label{tab:add-train-hyperparams-llava-ov}
\resizebox{0.9\columnwidth}{!}{%
{\renewcommand{\arraystretch}{1.0}
\begin{tabular}{ll}
Name                            & Value             \\ \midrule\midrule
bf16                            & TRUE              \\
tf32                            & True              \\
mm\_vision\_tower\_lr           & 2.00E-06          \\
mm\_projector\_type             & mlp2x\_gelu       \\
mm\_vision\_select\_layer       & -2                \\
image\_aspect\_ratio            & anyres\_max\_9    \\
image\_grid\_pinpoints          & "(1x1),...,(6x6)" \\
mm\_patch\_merge\_type          & spatial\_unpad    \\
per\_device\_train\_batch\_size & 1                 \\
per\_device\_eval\_batch\_size  & 1                 \\
gradient\_accumulation\_steps   & 16                \\
weight\_decay                   & 0                 \\
warmup\_ratio                   & 0.03              \\
lr\_scheduler\_type             & cosine            \\
model\_max\_length              & 8192  \\ \bottomrule           
\end{tabular}
}
}
\vspace{-5pt}
\end{table}
\section{Additional evaluation details}
\label{app:add_eval_details}
We present additional inference and evaluation details in Table~\ref{tab:add-eval-details} and the LLM judge prompts used for MMVet and MathVista from VLMEvalKit \cite{duan2024vlmevalkit} in Listings~\ref{lst:mmvet-judge} and \ref{lst:mathvista-judge}.

\section{License}
The datasets and models are for research purposes only in support of an academic paper. All artifacts from this paper are licensed under the noncommercial license CC-BY-NC 4.0. Some of our models were built with Meta Llama 3, which is licensed under the Meta Llama 3 Community License, Copyright © Meta Platforms, Inc. All Rights Reserved.

\begin{table*}[ht]
\centering
\caption{\textbf{Additional Results on Model-generated data ablations with Mantis-CLIP.} We observe similar results of data ablations on Mantis-CLIP as on Mantis-SigLIP.}
\label{tab:model-data-ablations-clip}
\vspace{-5pt}
\resizebox{\textwidth}{!}{%
{\renewcommand{\arraystretch}{1.1}
\begin{tabular}{llllrrrrrrrrrr}
Data source                             & Final data format                                                 & Size                  & Model                     & A-OKVQA & BLINK & MathVista & MMMU & MMStar & MMVet & MMVP & RealWorldQA & Avg  & Delta \\ \midrule \midrule
\multirow{6}{*}{All datasets}           & \da                                                & \multirow{2}{*}{293K} & Mantis-CLIP               & 80.7    & 45.8  & 33.1      & 42.2 & 36.7   & 28.9  & 62.7 & \textbf{52.3}        & 47.8 &       \\
                                        & \cota                                              &                       & \modelname & 81.1    & \textbf{49.6}  & \textbf{36.6}      & 42.8 & \textbf{40.8}   & \textbf{45.2}  & \textbf{63.3} & 51.1        & \textbf{51.3} & \textbf{3.5}   \\ \cmidrule{2-14}
                                        & \da                                                & \multirow{2}{*}{580K} & Mantis-CLIP               & 82.0    & 47.2  & 31.5      & 40.6 & 38.3   & 31.4  & \textbf{63.3} & 49.0        & 47.9 &       \\
                                        & \cota + \cotht                      &                       & \modelname & \textbf{82.6}    & 47.7  & 35.9      & 38.3 & 39.6   & 43.4  & 61.0 & 51.1        & 49.9 & 2.0   \\ \cmidrule{2-14}
                                        & \da                                                & \multirow{2}{*}{815K} & Mantis-CLIP               & 81.0    & 47.3  & 32.9      & 47.3 & 38.1   & 29.1  & 63.0 & 51.0        & 48.7 &       \\
                                        & \cota + \cotht + \da &                       & \modelname & 81.3    & 43.1  & 32.7      & \textbf{48.0} & 37.7   & 31.0  & 58.7 & 47.6        & 47.5 & -1.2  \\ \midrule
\multirow{2}{*}{\modelname-useful datasets} & \da                                                & \multirow{2}{*}{566K} & Mantis-CLIP               & 79.1    & 44.1  & 30.9      & 39.4 & 35.5   & 30.1  & 62.7 & 49.8        & 46.5 &       \\
                                        & \cota + \cotht + \da &                       & \modelname & 80.7    & 41.6  & 31.5      & 41.0 & 38.2   & 36.1  & 59.0 & 48.2        & 47.0 & 0.6       \\ \bottomrule\bottomrule
\end{tabular}
}
}
\end{table*}
\begin{figure*}[t]
    \centering
    \includegraphics[width=0.8\textwidth]{images/additional_examples.pdf}
    \vspace{-10pt}
    \caption{We present additional successful examples of \modelname. }
    \label{fig:add-examples}
    \vspace{-10pt}
\end{figure*}

\begin{figure*}[t]
    \centering
    \includegraphics[width=0.8\textwidth]{images/additional_examples_2.pdf}
    \vspace{-10pt}
    \caption{We present additional successful examples of \modelname. }
    \label{fig:add-examples-2}
    \vspace{-10pt}
\end{figure*}

\begin{figure*}[t]
    \centering
    \includegraphics[width=0.8\textwidth]{images/failure_examples.pdf}
    \vspace{-10pt}
    \caption{We present additional failure cases of \modelname. }
    \label{fig:failure-examples}
    \vspace{-10pt}
\end{figure*}
\begin{table*}[ht]
\centering
\caption{\textbf{Detailed statistics of the \cota\ 293K dataset.}}
\label{tab:dataset-stats}
\resizebox{\textwidth}{!}{%
{\renewcommand{\arraystretch}{0.9}
\begin{tabular}{clrrrrr}
Original Dataset           & Subset               & Instance \# & Max Image \# & Avg Image \# & Max Turn \# & Avg Turn \# \\ \midrule \midrule
\multirow{19}{*}{Cauldron} & ai2d                 & 694               & 2            & 1            & 6           & 2           \\
                           & aokvqa               & 6322              & 5            & 1.1          & 7           & 2.1         \\
                           & chartqa              & 4726              & 2            & 1            & 10          & 2.1         \\
                           & cocoqa               & 13289             & 3            & 1.1          & 4           & 2           \\
                           & dvqa                 & 2158              & 2            & 1            & 7           & 2.5         \\
                           & iconqa               & 3791              & 3            & 1.1          & 5           & 2.2         \\
                           & infographic\_vqa     & 3822              & 3            & 1            & 9           & 2.3         \\
                           & mimic\_cgd           & 6899              & 6            & 2.1          & 7           & 2.8         \\
                           & nlvr2                & 9716              & 4            & 2.1          & 6           & 2.5         \\
                           & ocrvqa               & 22991             & 2            & 1            & 7           & 2           \\
                           & scienceqa            & 850               & 2            & 1            & 6           & 2.3         \\
                           & st\_vqa              & 11322             & 3            & 1            & 8           & 2           \\
                           & tabmwp               & 14548             & 1            & 1            & 10          & 2.5         \\
                           & tallyqa              & 16171             & 3            & 1.4          & 5           & 2.1         \\
                           & textvqa              & 15475             & 5            & 1            & 6           & 2.1         \\
                           & visual7w             & 4773              & 3            & 1.1          & 5           & 2.1         \\
                           & vqarad               & 115               & 2            & 1            & 4           & 2.2         \\
                           & vqav2                & 13394             & 5            & 1.2          & 6           & 2.1         \\
                           & vsr                  & 1864              & 2            & 1.2          & 4           & 2.1         \\ \midrule
\multirow{12}{*}{Mantis}   & birds-to-words       & 742               & 4            & 2            & 5           & 2.7         \\
                           & coinstruct           & 31773             & 8            & 2.3          & 8           & 2.2         \\
                           & contrastive\_caption & 4296              & 8            & 3.6          & 6           & 2           \\
                           & dreamsim             & 1738              & 3            & 3            & 3           & 2           \\
                           & iconqa               & 6660              & 7            & 2.6          & 6           & 2.2         \\
                           & imagecode            & 559               & 18           & 10.1         & 10          & 3.1         \\
                           & lrv\_multi           & 3401              & 9            & 3.3          & 6           & 2.2         \\
                           & multi\_vqa           & 2089              & 7            & 3.8          & 8           & 2.6         \\
                           & nlvr2                & 5436              & 4            & 2            & 5           & 2.5         \\
                           & spot-the-diff        & 2591              & 5            & 2.8          & 8           & 3           \\
                           & nextqa               & 3057              & 15           & 8.2          & 9           & 2.3         \\
                           & llava\_665k\_multi   & 77843             & 11           & 2.2          & 10          & 2.1         \\ \midrule
\multicolumn{2}{c}{Total}                         & 293105            & 18           & 2.2          & 10          & 2.3        \\ \bottomrule
\end{tabular}
}
}
\end{table*}
\begin{table*}[ht]
\centering
\caption{\textbf{Templates for programmatic data generation.}}
\label{tab:program-data-templates}
\resizebox{\textwidth}{!}{%
{\renewcommand{\arraystretch}{0.9}
\begin{tabular}{llll}
\# of input images & Capabilities                                               & Question Template                                     & Action Template                                                                                                                   \\ \midrule \midrule
\multirow{10}{*}{1}    & \multirow{3}{*}{Counting}                                  & How many \{object\} are there?                        & \multirow{8}{*}{LocalizeObjects}                                                                                                  \\
                       &                                                            & Among \{objects\}, which is the most frequent object? &                                                                                                                                   \\
                       &                                                            & Among \{objects\}, which object appears the least?    &                                                                                                                                   \\ \cmidrule{2-3}
                       & Counting, Attribute recognition                            & How many \{attribute\} \{object\} are there?          &                                                                                                                                   \\ \cmidrule{2-3}
                       & \multirow{4}{*}{2D spatial reasoning}                      & Among \{objects\}, which is on the most left side?    &                                                                                                                                   \\
                       &                                                            & Among \{objects\}, which is on the most right side?   &                                                                                                                                   \\
                       &                                                            & Among \{objects\}, which is on the most top side?     &                                                                                                                                   \\
                       &                                                            & Among \{objects\}, which is on the most bottom side?  &                                                                                                                                   \\ \cmidrule{2-4}
                       & \multirow{2}{*}{3D spatial reasoning}                      & Which of \{objects\} is closer?                       & \multirow{2}{*}{\begin{tabular}[c]{@{}l@{}}LocalizeObjects, EstimateRegionDepth x2\\ OR, EstimateObjectDepth x2\end{tabular}} \\
                       &                                                            & Which of \{objects\} is farther?                      &                                                                                                                                   \\ \midrule
\multirow{6}{*}{2-3}   & Multi-image understanding                                  & Which image has \{object\}?                           & \multirow{6}{*}{LocalizeObjects x N}                                                                                              \\
                       & Multi-image understanding, Counting                        & How many \{object\} are in in these images?           &                                                                                                                                   \\
                       & Multi-image understanding, Counting                        & Which image has most \{object\}?                      &                                                                                                                                   \\
                       & Multi-image understanding, Counting                        & Which image has least \{object\}?                     &                                                                                                                                   \\
                       & Multi-image understanding, Attribute recognition           & Which image has \{attribute\} \{object\}?             &                                                                                                                                   \\
                       & Multi-image understanding, Attribute recognition, Counting & How many \{attribute\} \{object\} in these images?    &   \\ \bottomrule                                                                                                                               
\end{tabular}
}
}
\end{table*}
\begin{table*}[h]
\centering
\caption{\textbf{Additional inference and evaluation details.}}
\label{tab:add-eval-details}
\resizebox{0.7\textwidth}{!}{%
{\renewcommand{\arraystretch}{1.0}
\begin{tabular}{lll}
Stage                       & Name                                                                           & Value              \\ \midrule \midrule
\multirow{4}{*}{Inference}  & do\_sample                                                                     & FALSE              \\
                            & temperature                                                                    & 0                  \\
                            & max\_new\_tokens                                                               & 2000               \\
                            & max\_consecutive\_auto\_reply                                                  & 10                 \\ \midrule
\multirow{4}{*}{Evaluation} & llm judge for multiple choice  \& yes/no questions            & gpt-3.5-turbo-0125 \\
                            & llm judge for short answer  questions (i.e. MMVet, MathVista) & gpt-4-1106-preview \\
                            & llm judge max\_new\_tokens                                                     & 2048               \\
                            & llm judge retry                                                                & 5     \\ \bottomrule           
\end{tabular}
}
}
\end{table*}
\onecolumn
\clearpage
\begin{lstlisting}[language=Python, caption=Python implementation of all actions, label=lst:actions]
class BaseAction:
    """
    This is the Action class for agent to use.
    Using this Action class to wrap APIs, tools, models as an Action of an agent
    """

    def __init__(
        self,
        id: int,
        description: str = "",
        args_spec: dict = {},
        rets_spec: dict = {},
        examples: List = []
    ) -> None:
        """
        the agent action should be connected with data and env
        Args:
            id: the id of the action
            description: the description of the action
            args_spec: the specification of the arguments
            rets_spec: the specification of the returns
            examples: a list of examples of the action
        """
        self.name = self.__class__.__name__
        self.id = id
        self.description = description
        self.args_spec = args_spec
        self.rets_spec = rets_spec
        self.examples = examples
        self.device = "cuda:0" if torch.cuda.is_available() else "cpu"

    def __call__(self, **kwargs) -> str:
        """
        implement the Action as
        """
        raise NotImplementedError


class OCR(BaseAction):
    def __init__(self, id) -> None:
        description = "Extract texts from an image or return an empty string if no text is in the image. Note that the texts extracted may be incorrect or in the wrong order. It should be used as a reference only."
        args_spec = {"image": "the image to extract texts from."}
        rets_spec = {"text": "the texts extracted from the image."}
        examples = [{"name": "OCR", "arguments": {"image": "image-0"}}]

        super().__init__(
            id=id, description=description, args_spec=args_spec, rets_spec=rets_spec, examples=examples
        )
        
    def __call__(self, image, tool_version=LATEST_GPT_MODEL_ID):
        if tool_version == "easyocr":
            import easyocr
            import io
            reader = easyocr.Reader(["en"])  # Load the OCR model into memory
            image = image_processing(image)
            if isinstance(image, str):
                # If image is a path, use it directly
                image_path_or_bytes = (
                    image if os.path.exists(image) else get_full_path_data(image)
                )
            else:
                # If image is an Image object, convert it to a bytes stream
                buffer = io.BytesIO()
                image.save(buffer, format="JPEG")
                buffer.seek(0)
                image_path_or_bytes = buffer

            result = reader.readtext(image_path_or_bytes)
            result_text = [text for _, text, _ in result]
            result_formatted = {"text": ", ".join(result_text)}
        else:
            from openai import OpenAI
            import base64
            client = OpenAI(api_key=os.getenv("OPENAI_API_KEY"))

            def encode_image(image_path):
                with open(image_path, "rb") as image_file:
                    return base64.b64encode(image_file.read()).decode('utf-8')

            image_path = image_processing(image, return_path=True)
            base64_image = encode_image(image_path)
            
            response = client.chat.completions.create(
                model=tool_version,
                messages=[
                    {
                        "role"   : "user",
                        "content": [
                            {"type": "text", "text": f"What are the texts in the image?"},
                            {
                                "type"     : "image_url",
                                "image_url": {
                                    "url": f"data:image/jpeg;base64,{base64_image}",
                                },
                            },
                        ],
                    }
                ],
                max_tokens=300,
            )
            result_formatted = {"text": response.choices[0].message.content}

        return result_formatted


class GetObjects(BaseAction):
    def __init__(self, id) -> None:
        description = "Using this function to get objects in an image."
        args_spec = {"image": "the image to get objects from."}
        rets_spec = {"objects": "the objects detected in the image."}
        examples = [{"name": "GetObjects", "arguments": {"image": "image-0"}}]

        super().__init__(
            id=id, description=description, args_spec=args_spec, rets_spec=rets_spec, examples=examples
        )
    
    def __call__(self, image, tool_version="https://huggingface.co/xinyu1205/recognize-anything-plus-model/resolve/main/ram_plus_swin_large_14m.pth?download=true"):
        from ram.models import ram_plus
        from ram import get_transform, inference_ram_openset as inference
        
        model_path_or_url = tool_version
        image_size = 384
        transform = get_transform(image_size=image_size)
        
        vit_size = "swin_l"
        # load model
        model = ram_plus(pretrained=model_path_or_url,
                        image_size=image_size,
                        vit=vit_size)
        model.eval()
        model = model.to(self.device)
        image = image_processing(image)
        image = transform(image).unsqueeze(0).to(self.device)
        tags = inference(image, model)
        objs = tags.split(" | ")
        return {"objects": objs}


class VisualizeRegionsOnImage(BaseAction):
    def __init__(self, id) -> None:
        description = "Using this function to label regions on an image."
        args_spec = {"image": "the image to label.", 
                     "regions": "the regions to label on the image, where each region is represented by a dictionary with the region's bounding box and label text (can be empty string).",
                     "color": "an optional argument that specifies the color of the bounding box."
                    }
        rets_spec = {"image": "the image with regions labeled."}
        examples = [
            {"name": "VisualizeRegionsOnImage", "arguments": {"image": "image-0", "regions": [{"label": "", "bbox": [0.3, 0.2, 0.5, 0.4]}]}},
            {"name": "VisualizeRegionsOnImage", "arguments": {"image": "image-0", "regions": [{"label": "cat", "bbox": [0.3, 0.2, 0.5, 0.4]}], "color": "red"}}
        ]

        super().__init__(
            id=id, description=description, args_spec=args_spec, rets_spec=rets_spec, examples=examples
        )
    
    def __call__(self, image, regions: List[Region], color='yellow', width=4):
        image = image_processing(image)
        text_color = 'black'
        W,H = image.size
        img1 = image.copy()
        draw = ImageDraw.Draw(img1)
        font = ImageFont.truetype('/usr/share/fonts/truetype/dejavu/DejaVuSansMono-Bold.ttf', 16)
        for i, obj in enumerate(regions):
            bbox = obj['bbox']
            bbox = bbox[0] * W, bbox[1] * H, bbox[2] * W, bbox[3] * H
            draw.rectangle(bbox, outline=color, width=width)
            x1, y1, x2, y2 = bbox
            label = obj['label'] if "label" in obj else ""
            w,h = font.getsize(label)
            if x1 + w > W or y2  +h > H:
                draw.rectangle((x1, y2 - h, x1 + w, y2), fill=color)
                draw.text((x1, y2-h),label,fill=text_color,font=font)
            else:
                draw.rectangle((x1, y2, x1 + w, y2 + h), fill=color)
                draw.text((x1, y2),label,fill=text_color,font=font)
        return {"image": img1}

    
class LocalizeObjects(BaseAction):
    def __init__(self, id) -> None:
        description = "Localize one or multiple objects/regions with bounding boxes. This tool may output objects that don't exist or miss objects that do. You should use the output only as weak evidence for reference. When answering questions about the image, you should double-check the detected objects. You should be especially cautious about the total number of regions detected, which can be more or less than the actual number."
        args_spec = {
            "image": "the image to localize objects/regions in.", 
            "objects": "a list of object names to localize. e.g. ['dog', 'cat', 'person']. the model might not be able to detect rare objects or objects with complex descriptionriptions."
        }
        rets_spec = {"image": "the image with objects localized and visualized on it.", "regions": "the regions of interests localized in the image, where each region is represented by a dictionary with the region's label text, bounding box and confidence score. The confidence score is between 0 and 1, where 1 means the model is very confident. Note that both the bounding boxes and confidence scores can be unreliable and should only be used as reference."}
        examples = [{"name": "LocalizeObjects", "arguments": {"image": "image-0", "objects": ["dog", "cat"]}}]
        
        super().__init__(
            id=id, description=description, args_spec=args_spec, rets_spec=rets_spec, examples=examples
        )
    
    def __call__(self, image, objects: List[str]):
        from groundingdino.util.inference import load_model, load_image, predict, annotate
        import cv2
        text = ". ".join(objects)
        model = load_model("/user/mma/mma/GroundingDINO/groundingdino/config/GroundingDINO_SwinT_OGC.py", 
                           "/user/mma/mma/GroundingDINO/weights/groundingdino_swint_ogc.pth",
                           device=self.device)
        BOX_TRESHOLD = 0.35
        TEXT_TRESHOLD = 0.25
        image_path = image_processing(image, return_path=True)
        original_image = image_processing(image)
        image_source, image = load_image(image_path)

        boxes, logits, phrases = predict(
            model=model,
            image=image,
            caption=text,
            box_threshold=BOX_TRESHOLD,
            text_threshold=TEXT_TRESHOLD
        )

        objects = []
        obj_cnt = {}
        for i in range(len(boxes)):
            xyxy = box_convert(boxes=boxes[i], in_fmt="cxcywh", out_fmt="xyxy").numpy()
            bbox = [round(val, 2) for val in list(xyxy)]
            score = round(logits[i].item(), 2)
            phrase = phrases[i]
            obj_cnt[phrase] = obj_cnt.get(phrase, 0) + 1
            phrase = f"{phrase}-{obj_cnt[phrase]}" if obj_cnt[phrase] > 1 else phrase
            objects.append({"label": phrase, "bbox": bbox, "score": score})
        visualize = VisualizeRegionsOnImage(0)
        results = visualize(image=original_image, regions=objects)
        tagged_image = results["image"]
        results_formatted = {"regions": objects, "image": tagged_image}
        return results_formatted


class Crop(BaseAction):
    def __init__(self, id) -> None:
        description = "Crop an image with the bounding box. It labels the cropped region with a bounding box and crops the region with some margins around the bounding box to help with contextual understanding of the region."
        args_spec = {
            "image": "the image to crop.",
            "bbox": "the bbox to crop. It should be a list of [left, top, right, bottom], where each value is a float between 0 and 1 to represent the percentage of the image width/height and how far it is from the top left corner at [0, 0].",
        }
        rets_spec = {"image": "the cropped image."}
        examples = [{"name": "Crop", "arguments": {"image": "image-0", "bbox": [0.33, 0.21, 0.58, 0.46]}}]

        super().__init__(
            id=id, description=description, args_spec=args_spec, rets_spec=rets_spec, examples=examples
        )
    
    def __call__(self, image, bbox):
        image = image_processing(image)

        if isinstance(bbox, str):
            try:
                bbox = ast.literal_eval(bbox)
            except:
                bbox = []

        use_percent = (all(x <= 1.0 for x in bbox))
        if not use_percent:
            raise ValueError("Bounding box coordinates must be between 0 and 1.")
        
        visualize = VisualizeRegionsOnImage(0)
        results = visualize(image=image, regions=[{"label": "", "bbox": bbox}])
        image = results["image"]
        
        W, H = image.size
        bbox = [bbox[0] * W, bbox[1] * H, bbox[2] * W, bbox[3] * H]
        bbox = expand_bbox(bbox, image.size)
        out_img = image.crop(bbox)
        return {"image": out_img}
    

class ZoomIn(BaseAction):
    def __init__(self, id) -> None:
        description = "Zoom in on a region of the input image. This tool first crops the specified region from the image with the bounding box and then resizes the cropped region to create the zoom effect. It also adds some margins around the cropped region to help with contextual understanding of the region."
        args_spec = {
            "image": "the image to zoom in on.",
            "bbox": "The bbox should be a list of [left, top, right, bottom], where each value is a float between 0 and 1 to represent the percentage of the image width/height and how far it is from the top left corner at [0, 0].",
            "zoom_factor": "the factor to zoom in by. It should be greater than 1.",
        }
        rets_spec = {"image": "the zoomed in image."}
        examples = [
            {"name": "ZoomIn", "arguments": {"image": "image-0", "bbox": [0.4, 0.3, 0.5, 0.4], "zoom_factor": 2}},
        ]

        super().__init__(
            id=id, description=description, args_spec=args_spec, rets_spec=rets_spec, examples=examples
        )
    
    def __call__(self, image, bbox, zoom_factor):
        if zoom_factor <= 1:
            raise ValueError("Zoom factor must be greater than 1 to zoom in")

        image = image_processing(image)
        use_percent = (all(x <= 1.0 for x in bbox))
        if not use_percent:
            raise ValueError("Bounding box coordinates must be between 0 and 1.")
        
        crop = Crop(0)
        cropped_image = crop(image, bbox)["image"]
       
        W, H = cropped_image.size
        
        # Calculate the size of the zoomed image
        new_width = int(W * zoom_factor)
        new_height = int(H * zoom_factor)
        
        # Resize the cropped image to create the zoom effect
        zoomed_image = cropped_image.resize((new_width, new_height), Image.LANCZOS)
        return {'image': zoomed_image}


class GetImageToImagesSimilarity(BaseAction):
    def __init__(self, id) -> None:
        description = "Get the similarity between one image and a list of other images. Note that this similarity score may not be accurate and should be used as a reference only."
        args_spec = {
            "image": "the reference image.",
            "other_images": "the other images to compare to the reference image.",
        }
        rets_spec = {"similarity": "the CLIP similarity scores between the reference image and the other images.", "best_image_index": "the index of the most similar image."}
        examples = [
            {"name": "GetImageToImagesSimilarity", "arguments": {"image": "image-0", "other_images": ["image-1", "image-2"]}}
        ]

        super().__init__(
            id=id, description=description, args_spec=args_spec, rets_spec=rets_spec, examples=examples
        )

    def __call__(self, image, other_images, tool_version='ViT-H-14-378-quickgelu', other_images_raw=None):
        import torch
        import open_clip
        original_images = other_images_raw
        model, _, preprocess = open_clip.create_model_and_transforms(tool_version, pretrained='dfn5b')
        model.eval()
        image = image_processing(image)
        images = [image_processing(image) for image in other_images]

        image = preprocess(image).unsqueeze(0)
        images = torch.stack([preprocess(image) for image in images])

        with torch.no_grad(), torch.cuda.amp.autocast():
            image1_features = model.encode_image(image)
            image2_features = model.encode_image(images)

            image1_features /= image1_features.norm(dim=-1, keepdim=True)
            image2_features /= image2_features.norm(dim=-1, keepdim=True)

            probs = image1_features @ image2_features.T
        sim_scores = [round(sim_score, 2) for sim_score in probs[0].tolist()]
        best_image_match = torch.argmax(probs).item()
        return {'similarity': sim_scores, "best_image_index": best_image_match, "best_image": original_images[best_image_match]}


class GetImageToTextsSimilarity(BaseAction):
    def __init__(self, id) -> None:
        description = "Get the similarity between one image and a list of texts. Note that this similarity score may not be accurate and should be used as a reference only."
        args_spec = {
            "image": "the reference image.",
            "texts": "a list of texts to compare to the reference image.",
        }
        rets_spec = {"similarity": "the CLIP similarity between the image and the texts.", "best_text_index": "the index of the most similar text.", "best_text": "the most similar text."}
        examples = [
            {"name": "GetImageToTextsSimilarity", "arguments": {"image": "image-0", "texts": ["a cat", "a dog"]}}
        ]

        super().__init__(
            id=id, description=description, args_spec=args_spec, rets_spec=rets_spec, examples=examples
        )
    
    def __call__(self, image, texts, tool_version='ViT-H-14-378-quickgelu'):
        import torch
        import open_clip

        model, _, preprocess = open_clip.create_model_and_transforms(tool_version, pretrained='dfn5b')
        model.eval()  # model in train mode by default, impacts some models with BatchNorm or stochastic depth active
        tokenizer = open_clip.get_tokenizer(tool_version)

        image = preprocess(image_processing(image)).unsqueeze(0)
        text = tokenizer(texts)

        with torch.no_grad(), torch.cuda.amp.autocast():
            image_features = model.encode_image(image)
            text_features = model.encode_text(text)
            image_features /= image_features.norm(dim=-1, keepdim=True)
            text_features /= text_features.norm(dim=-1, keepdim=True)

            probs = image_features @ text_features.T
        sim_scores = [round(sim_score, 2) for sim_score in probs[0].tolist()]
        best_text_match = torch.argmax(probs).item()
        return {'similarity': sim_scores, "best_text_index": best_text_match, "best_text": texts[best_text_match]}


class GetTextToImagesSimilarity(BaseAction):
    def __init__(self, id) -> None:
        description = "Get the similarity between one text and a list of images. Note that this similarity score may not be accurate and should be used as a reference only."
        args_spec = {
            "text": "the reference text.",
            "images": "a list of images to compare to the reference text.",
        }
        rets_spec = {"similarity": "the CLIP similarity between the image and the texts.", "best_image_index": "the index of the most similar image."}
        examples = [
            {"name": "GetTextToImagesSimilarity", "arguments": {"text": "a black and white cat", "images": ["image-0", "image-1"]}}
        ]
        
        super().__init__(
            id=id, description=description, args_spec=args_spec, rets_spec=rets_spec, examples=examples
        )
    
    def __call__(self, text, images, tool_version='ViT-H-14-378-quickgelu'):
        import torch
        import open_clip
        original_images = images
        model, _, preprocess = open_clip.create_model_and_transforms(tool_version, pretrained='dfn5b')
        model.eval()  # model in train mode by default, impacts some models with BatchNorm or stochastic depth active
        tokenizer = open_clip.get_tokenizer(tool_version)

        text = tokenizer([text])
        images = [image_processing(image) for image in images]
        images = torch.stack([preprocess(image) for image in images])

        with torch.no_grad(), torch.cuda.amp.autocast():
            image_features = model.encode_image(images)
            text_features = model.encode_text(text)
            image_features /= image_features.norm(dim=-1, keepdim=True)
            text_features /= text_features.norm(dim=-1, keepdim=True)

            probs = text_features @ image_features.T
        sim_scores = [round(sim_score, 2) for sim_score in probs[0].tolist()]
        best_image_match = torch.argmax(probs).item()
        return {'similarity': sim_scores, "best_image_index": best_image_match, "best_image": original_images[best_image_match]} 


class EstimateObjectDepth(BaseAction):
    def __init__(self, id) -> None:
        description = "Estimate the depth of an object in an image using DepthAnything model. It returns an estimated depth value of the object specified by the a brief text description. The smaller the value is, the closer the object is to the camera, and the larger the farther. This tool may help you to better reason about the spatial relationship, like which object is closer to the camera."
        args_spec = {
            "image": "the image to get the depth from.",
            "object": "a short description of the object to get the depth from.",
        }
        rets_spec = {"depth": "the estimated depth of the object."}
        examples = [
            {"name": "EstimateObjectDepth", "arguments": {"image": "image-0", "object": "a black cat"}},
        ]

        super().__init__(
            id=id, description=description, args_spec=args_spec, rets_spec=rets_spec, examples=examples
        )
    
    def __call__(self, image, object, mode="mean"):
        action = LocalizeObjects(0)
        results = action(image=image, objects=[object])
        if len(results["regions"]) == 0:
            return {"depth": "Object not found."}
        else:
            # use the best match object's bbox
            best_match = np.argmax([region["score"] for region in results["regions"]])
            bbox = results["regions"][best_match]["bbox"]
            depth_estimator = EstimateRegionDepth(0)
            return depth_estimator(image=image, bbox=bbox, mode=mode)
        

class EstimateRegionDepth(BaseAction):
    def __init__(self, id) -> None:
        description = "Estimate the depth of a region in an image using DepthAnything model. It returns an estimated depth value of the region specified by the input bounding box. The smaller the value is, the closer the region is to the camera, and the larger the farther. This tool may help you to better reason about the spatial relationship, like which object is closer to the camera. "
        args_spec = {
            "image": "the image to get the depth from.",
            "bbox": "the bbox of the region to get the depth from. It should be a list of [left, top, right, bottom], where each value is a float between 0 and 1 to represent the percentage of the image width/height and how far it is from the top left corner at [0, 0].",
            # "mode": "the mode to get the depth. It should be one of 'center' or 'average'. 'center' returns the depth of the center of the region. 'average' returns the average depth of the region.",
        }
        rets_spec = {"depth": "the estimated depth of the region."}
        examples = [
            {"name": "EstimateRegionDepth", "arguments": {"image": "image-0", "bbox": [0.3, 0.2, 0.5, 0.4]}},
        ]
        super().__init__(
            id=id, description=description, args_spec=args_spec, rets_spec=rets_spec, examples=examples
        )
        
    def __call__(self, image, bbox: List[str], mode="mean"):
        import numpy as np
        from scipy import stats
        image = image_processing(image)
        depth_model = pipeline(task="depth-estimation", model="depth-anything/Depth-Anything-V2-Small-hf", device=self.device)
        result = depth_model(image)
        depth = result["predicted_depth"][0].numpy()
        depth = depth.max() - depth # smaller values in depth map are farther from the camera so reversing the values
        H, W = depth.shape
   
        use_percent = all(x <= 1.0 for x in bbox)
        if not use_percent:
            raise ValueError("Bounding box coordinates must be between 0 and 1.")
        bbox = [bbox[0] * W, bbox[1] * H, bbox[2] * W, bbox[3] * H]
        if mode == "center":
            x, y = (bbox[0] + bbox[2]) / 2, (bbox[1] + bbox[3]) / 2
            x, y = int(x), int(y)
            depth_value = depth[y, x]
        elif mode == "mean":
            x1, y1, x2, y2 = bbox
            x1, y1, x2, y2 = int(x1), int(y1), int(x2), int(y2)
            depth_value = np.mean(depth[y1:y2, x1:x2])
        elif mode == "mode":
            x1, y1, x2, y2 = bbox
            x1, y1, x2, y2 = int(x1), int(y1), int(x2), int(y2)
            mode_result = stats.mode(depth[y1:y2, x1:x2])
            depth_value = mode_result.mode[0]
        else:
            raise NotImplementedError(f"Depth mode {mode} is not supported.")
        return {"depth": round(depth_value, 2)}


class Calculate(BaseAction):
    def __init__(self, id) -> None:
        description = "Calculate a math expression."
        args_spec = {"expression": "the math expression to calculate."}
        rets_spec = {"result": "the result of the math expression."}
        examples = [
            {"name": "Calculate", "arguments": {"expression": "2 + 2"}}, 
            {"name": "Calculate", "arguments": {"expression": "4*9*84"}},
            {"name": "Calculate", "arguments": {"expression": "5-4/2"}},
        ]

        super().__init__(
            id=id, description=description, args_spec=args_spec, rets_spec=rets_spec, examples=examples
        )
    
    def __call__(self, expression):
        result = eval(expression)
        return {"result": result}
        
    
class SolveMathEquation(BaseAction):
    def __init__(self, id) -> None:
        description = "Using this action to solve a math problem with WolframAlpha."
        args_spec = {"query": "a question that involves a math equation to be solved."}
        rets_spec = {"result": "the result of the query."}
        examples = [
            {"name": "SolveMathEquation", "arguments": {"query": "2 + 2=?"}},
            {"name": "SolveMathEquation", "arguments": {"query": "x^2 + 2x + 1 = 0, what is x?"}},
        ]
        
        self.client = wolframalpha.Client(os.getenv("WOLFRAM_ALPHA_API_KEY"))
        super().__init__(
            id=id, description=description, args_spec=args_spec, rets_spec=rets_spec, examples=examples
        )
        
    def __call__(self, query):
        from urllib.error import HTTPError

        is_success = False  
       
        res = self.client.query(query)

        if not res["@success"]:
            return (
                "Your Wolfram query is invalid. Please try a new query for wolfram.",
                is_success,
            )
        assumption = next(res.pods).text
        answer = ""
        for result in res["pod"]:
            if result["@title"] == "Solution":
                answer = result["subpod"]["plaintext"]
            if result["@title"] == "Results" or result["@title"] == "Solutions":
                for i, sub in enumerate(result["subpod"]):
                    answer += f"ans {i}: " + sub["plaintext"] + "\n"
                break
        if answer == "":
            answer = next(res.results).text

        if answer is None or answer == "":
            return {"result": "No good Wolfram Alpha Result was found"}
        else:
            return {"result": answer} 


class DetectFaces(BaseAction):
    def __init__(self, id) -> None:
        description = "Using this function to detect faces in an image."
        args_spec = {"image": "the image to detect faces from."}
        rets_spec = {"image": "the image with objects localized and visualized on it.", "regions": "the regions of the faces detected, where each regin is represented by a dictionary with the region's label text and bounding box."}
        examples = [
            {"name": "DetectFaces", "arguments": {"image": "image-0"}}
        ]
        import face_detection
        ckpt_path = f"/root/.cache/torch/hub/checkpoints/WIDERFace_DSFD_RES152.pth"
        if not os.path.exists(ckpt_path):
            from huggingface_hub import hf_hub_download
            hf_hub_download(repo_id="user/mma", filename="WIDERFace_DSFD_RES152.pth", local_dir="/root/.cache/torch/hub/checkpoints/")

        self.model = face_detection.build_detector(
            "DSFDDetector", confidence_threshold=.5, nms_iou_threshold=.3)
        super().__init__(
            id=id, description=description, args_spec=args_spec, rets_spec=rets_spec, examples=examples
        )

    def enlarge_face(self,box,W,H,f=1.5):
        x1,y1,x2,y2 = box
        w = int((f-1)*(x2-x1)/2)
        h = int((f-1)*(y2-y1)/2)
        x1 = max(0,x1-w)
        y1 = max(0,y1-h)
        x2 = min(W,x2+w)
        y2 = min(H,y2+h)
        return [x1,y1,x2,y2]
    
    def __call__(self, image):
        import numpy as np
        image = image_processing(image)

        with torch.no_grad():
            faces = self.model.detect(np.array(image))
        
        W,H = image.size
        objs = []
        for i,box in enumerate(faces):
            x1,y1,x2,y2,c = [int(v) for v in box.tolist()]
            normalized_bbox = [x1/W, y1/H, x2/W, y2/H]
            objs.append(dict(
                bbox=[round(num, 2) for num in normalized_bbox],
                label=f'face {i+1}' if i > 0 else 'face',
            ))
        visualize = VisualizeRegionsOnImage(0)
        results = visualize(image=image, regions=objs)
        tagged_image = results["image"]
        results_formatted = {"regions": objs, "image": tagged_image}
        return results_formatted


class QueryLanguageModel(BaseAction):
    def __init__(self, id) -> None:
        description = "Using this function to ask a language model a question."
        args_spec = {"query": "the question to ask the language model."}
        rets_spec = {"result": "the response from the language model."}
        examples = [
            {"name": "QueryLanguageModel", "arguments": {"query": "What is the capital of France?"}},
        ]
        super().__init__(
            id=id, description=description, args_spec=args_spec, rets_spec=rets_spec, examples=examples
        )
    
    def __call__(self, query):
        from openai import OpenAI
        client = OpenAI(api_key=os.getenv("OPENAI_API_KEY"))

        response = client.chat.completions.create(
            model=LATEST_GPT_MODEL_ID, 
            messages=[
                {
                    "role"   : "user",
                    "content": [
                        {"type": "text", "text": f"{query}"},
                    ],
                }
            ],
            max_tokens=300,
        )

        return {'result': response.choices[0].message.content}


class QueryKnowledgeBase(BaseAction):
    def __init__(self, id) -> None:
        description = "Using this function to query a knowledge base."
        args_spec = {"query": "the query to search in a knowledge base such as wikipedia."}
        rets_spec = {"result": "the answer from the knowledge base."}
        examples = [
            {"name": "QueryKnowledgeBase", "arguments": {"query": "Paris"}},
        ]

        super().__init__(
            id=id, description=description, args_spec=args_spec, rets_spec=rets_spec, examples=examples
        )
    
    def __call__(self, query, lang="en", sentences=2, knowledge_base="wikipedia"):
        if knowledge_base == "wikipedia":
            # Set the language for Wikipedia (default is 'en' for English)
            wikipedia.set_lang(lang)
            
            # Search Wikipedia for pages related to the query
            search_results = wikipedia.search(query)
            if not search_results:
                return {"No results found."}

            # Get the summary of the first search result
            page = wikipedia.page(search_results[0])
            summary = wikipedia.summary(page.title, sentences=sentences)
            result = {
                "title": page.title,
                "url": page.url,
                "summary": summary
            }
            return result
        else:
            raise NotImplementedError(f"Knowledge base {knowledge_base} is not supported.")


class Terminate(BaseAction):
    def __init__(self, id) -> None:
        description = "Using this function to finish the task."
        args_spec = {"answer": "the final answer."}
        rets_spec = {"answer": "the final answer."}
        examples = [{"name": "Terminate", "arguments": {"answer": "yes"}}]

        super().__init__(
            id=id, description=description, args_spec=args_spec, rets_spec=rets_spec, examples=examples
        )
    
    def __call__(self, answer):
        return {"answer": answer}
\end{lstlisting}
\twocolumn
\onecolumn
\begin{lstlisting}[caption=Model-based data generation system prompt, label=lst:data-gen-prompt]
[BEGIN OF GOAL]
You are a helpful assistant, and your goal is to solve the # USER REQUEST #. You can either rely on your own capabilities or perform actions with external tools to help you. A list of all available actions are provided to you in the below.
[END OF GOAL]

[BEGIN OF ACTIONS]
Name: OCR
Description: Extract texts from an image or return an empty string if no text is in the image. Note that the texts extracted may be incorrect or in the wrong order. It should be used as a reference only.
Arguments: {'image': 'the image to extract texts from.'}
Returns: {'text': 'the texts extracted from the image.'}
Examples:
{"name": "OCR", "arguments": {"image": "image-0"}}

Name: LocalizeObjects
Description: Localize one or multiple objects/regions with bounding boxes. This tool may output objects that don't exist or miss objects that do. You should use the output only as weak evidence for reference. When answering questions about the image, you should double-check the detected objects. You should be especially cautious about the total number of regions detected, which can be more or less than the actual number.
Arguments: {'image': 'the image to localize objects/regions in.', 'objects': "a list of object names to localize. e.g. ['dog', 'cat', 'person']. the model might not be able to detect rare objects or objects with complex descriptionriptions."}
Returns: {'image': 'the image with objects localized and visualized on it.', 'regions': "the regions of interests localized in the image, where each region is represented by a dictionary with the region's label text, bounding box and confidence score. The confidence score is between 0 and 1, where 1 means the model is very confident. Note that both the bounding boxes and confidence scores can be unreliable and should only be used as reference."}
Examples:
{"name": "LocalizeObjects", "arguments": {"image": "image-0", "objects": ["dog", "cat"]}}

Name: GetObjects
Description: Using this function to get objects in an image.
Arguments: {'image': 'the image to get objects from.'}
Returns: {'objects': 'the objects detected in the image.'}
Examples:
{"name": "GetObjects", "arguments": {"image": "image-0"}}

Name: EstimateRegionDepth
Description: Estimate the depth of a region in an image using DepthAnything model. It returns an estimated depth value of the region specified by the input bounding box. The smaller the value is, the closer the region is to the camera, and the larger the farther. This tool may help you to better reason about the spatial relationship, like which object is closer to the camera. 
Arguments: {'image': 'the image to get the depth from.', 'bbox': 'the bbox of the region to get the depth from. It should be a list of [left, top, right, bottom], where each value is a float between 0 and 1 to represent the percentage of the image width/height and how far it is from the top left corner at [0, 0].'}
Returns: {'depth': 'the estimated depth of the region.'}
Examples:
{"name": "EstimateRegionDepth", "arguments": {"image": "image-0", "bbox": [0.3, 0.2, 0.5, 0.4]}}

Name: EstimateObjectDepth
Description: Estimate the depth of an object in an image using DepthAnything model. It returns an estimated depth value of the object specified by the a brief text description. The smaller the value is, the closer the object is to the camera, and the larger the farther. This tool may help you to better reason about the spatial relationship, like which object is closer to the camera.
Arguments: {'image': 'the image to get the depth from.', 'object': 'a short description of the object to get the depth from.'}
Returns: {'depth': 'the estimated depth of the object.'}
Examples:
{"name": "EstimateObjectDepth", "arguments": {"image": "image-0", "object": "a black cat"}}

Name: Crop
Description: Crop an image with the bounding box. It labels the cropped region with a bounding box and crops the region with some margins around the bounding box to help with contextual understanding of the region.
Arguments: {'image': 'the image to crop.', 'bbox': 'the bbox to crop. It should be a list of [left, top, right, bottom], where each value is a float between 0 and 1 to represent the percentage of the image width/height and how far it is from the top left corner at [0, 0].'}
Returns: {'image': 'the cropped image.'}
Examples:
{"name": "Crop", "arguments": {"image": "image-0", "bbox": [0.33, 0.21, 0.58, 0.46]}}

Name: ZoomIn
Description: Zoom in on a region of the input image. This tool first crops the specified region from the image with the bounding box and then resizes the cropped region to create the zoom effect. It also adds some margins around the cropped region to help with contextual understanding of the region.
Arguments: {'image': 'the image to zoom in on.', 'bbox': 'The bbox should be a list of [left, top, right, bottom], where each value is a float between 0 and 1 to represent the percentage of the image width/height and how far it is from the top left corner at [0, 0].', 'zoom_factor': 'the factor to zoom in by. It should be greater than 1.'}
Returns: {'image': 'the zoomed in image.'}
Examples:
{"name": "ZoomIn", "arguments": {"image": "image-0", "bbox": [0.4, 0.3, 0.5, 0.4], "zoom_factor": 2}}

Name: QueryLanguageModel
Description: Using this function to ask a language model a question.
Arguments: {'query': 'the question to ask the language model.'}
Returns: {'result': 'the response from the language model.'}
Examples:
{"name": "QueryLanguageModel", "arguments": {"query": "What is the capital of France?"}}

Name: GetImageToImagesSimilarity
Description: Get the similarity between one image and a list of other images. Note that this similarity score may not be accurate and should be used as a reference only.
Arguments: {'image': 'the reference image.', 'other_images': 'the other images to compare to the reference image.'}
Returns: {'similarity': 'the CLIP similarity scores between the reference image and the other images.', 'best_image_index': 'the index of the most similar image.'}
Examples:
{"name": "GetImageToImagesSimilarity", "arguments": {"image": "image-0", "other_images": ["image-1", "image-2"]}}

Name: GetImageToTextsSimilarity
Description: Get the similarity between one image and a list of texts. Note that this similarity score may not be accurate and should be used as a reference only.
Arguments: {'image': 'the reference image.', 'texts': 'a list of texts to compare to the reference image.'}
Returns: {'similarity': 'the CLIP similarity between the image and the texts.', 'best_text_index': 'the index of the most similar text.', 'best_text': 'the most similar text.'}
Examples:
{"name": "GetImageToTextsSimilarity", "arguments": {"image": "image-0", "texts": ["a cat", "a dog"]}}

Name: GetTextToImagesSimilarity
Description: Get the similarity between one text and a list of images. Note that this similarity score may not be accurate and should be used as a reference only.
Arguments: {'text': 'the reference text.', 'images': 'a list of images to compare to the reference text.'}
Returns: {'similarity': 'the CLIP similarity between the image and the texts.', 'best_image_index': 'the index of the most similar image.'}
Examples:
{"name": "GetTextToImagesSimilarity", "arguments": {"text": "a black and white cat", "images": ["image-0", "image-1"]}}

Name: DetectFaces
Description: Using this function to detect faces in an image.
Arguments: {'image': 'the image to detect faces from.'}
Returns: {'image': 'the image with objects localized and visualized on it.', 'regions': "the regions of the faces detected, where each regin is represented by a dictionary with the region's label text and bounding box."}
Examples:
{"name": "DetectFaces", "arguments": {"image": "image-0"}}

Name: QueryKnowledgeBase
Description: Using this function to query a knowledge base.
Arguments: {'query': 'the query to search in a knowledge base such as wikipedia.'}
Returns: {'result': 'the answer from the knowledge base.'}
Examples:
{"name": "QueryKnowledgeBase", "arguments": {"query": "Paris"}}

Name: Calculate
Description: Calculate a math expression.
Arguments: {'expression': 'the math expression to calculate.'}
Returns: {'result': 'the result of the math expression.'}
Examples:
{"name": "Calculate", "arguments": {"expression": "2 + 2"}}
{"name": "Calculate", "arguments": {"expression": "4*9*84"}}
{"name": "Calculate", "arguments": {"expression": "5-4/2"}}

Name: SolveMathEquation
Description: Using this action to solve a math problem with WolframAlpha.
Arguments: {'query': 'a question that involves a math equation to be solved.'}
Returns: {'result': 'the result of the query.'}
Examples:
{"name": "SolveMathEquation", "arguments": {"query": "2 + 2=?"}}
{"name": "SolveMathEquation", "arguments": {"query": "x^2 + 2x + 1 = 0, what is x?"}}

Name: Terminate
Description: Using this function to finish the task.
Arguments: {'answer': 'the final answer.'}
Returns: {'answer': 'the final answer.'}
Examples:
{"name": "Terminate", "arguments": {"answer": "yes"}}

[END OF ACTIONS]

[BEGIN OF TASK INSTRUCTIONS]
1. You must only select actions from # ACTIONS #.
2. You can only call one action at a time.
3. If no action is needed, please make actions an empty list (i.e. ''actions'': []).
4. You must always call Terminate with your final answer at the end.
[END OF TASK INSTRUCTIONS]

[BEGIN OF FORMAT INSTRUCTIONS]
Your output should be in a strict JSON format as follows:
{"thought": "the thought process, or an empty string", "actions": [{"name": "action1", "arguments": {"argument1": "value1", "argument2": "value2"}}]}
[END OF FORMAT INSTRUCTIONS]

[BEGIN OF EXAMPLES]:
# USER REQUEST #:
 In image-0, Which of the two objects on the plate is the biggest?
A. The pile of scrambled eggs is the biggest.
B. The strawberries are the biggest object.
Please answer directly with only the letter of the correct option and nothing else.
# RESPONSE #:
{"thought": "To determine which of the two objects on the plate is larger, I need to analyze the size of the scrambled eggs, and the strawberries", "actions": [{"name": "LocalizeObjects", "arguments": {"image": "image-0", "objects": ["scrambled eggs", "strawberries"]}}]}
OBSERVATION:
{"image": "image-1", "regions": [{"label": "eggs", "bbox": [0.5, 0.6, 0.6, 0.8], "score": 0.85}, {"label": "strawberries", "bbox": [0.4, 0.5, 0.45, 0.7], "score": 0.54}]}
{"thought": "To calculate the area of a bounding box, we can use the formula: area = (x_max - x_min) * (y_max - y_min). We first get the area of the scrambled eggs.", "actions": [{"name": "Calculate", "arguments": {"expression": "(0.6-0.5) * (0.8-0.6)"}}]}
OBSERVATION:
{"result": "0.02"}
{"thought": "Then, we also calculate the area of the strawberries.", "actions": [{"name": "Calculate", "arguments": {"expression": "(0.45-0.4) * (0.7-0.5)"}}]}
OBSERVATION:
{"result": "0.01"}
{"thought": "Since 0.02 > 0.01, it is apparent that the eggs cover a larger area within their bounding box.", "actions": [{"name": "Terminate", "arguments": {"answer": "A"}}]}

# USER REQUEST #:
 Given the input image image-0, How many pedestrians are there in the image? Please answer directly with a single word or number.
# RESPONSE #:
{"thought": "To determine the number of pedestrians, I need to first localize them on the image.", "actions": [{"name": "LocalizeObjects", "arguments": {"image": "image-0", "objects": ["person"]}}]}
OBSERVATION:
{"image": "image-1", "regions": [{"label": "person", "bbox": [0.77, 0.47, 0.79, 0.54], "score": 0.83}, {"label": "person-2", "bbox": [0.69, 0.49, 0.7, 0.52], "score": 0.43}]}
{"thought": "The LocalizeObjects action returns two regions for \"person\", but one of the regions has a lower confidence score. Upon a closer look at the output image image-1, we can see that there is actually only one pedestrian in the image.", "actions": [{"name": "Terminate", "arguments": {"answer": "1"}}]}

# USER REQUEST #:
 Based on image-0, is the object on top bigger than the object below?
A. The object on the bottom is bigger.
B. The object on top is bigger.
C. Both objects are the same size.
Please answer directly with only the letter of the correct option and nothing else.
# RESPONSE #:
{"thought": "By looking at the image, we can see that both objects are game consoles of the same brand and size.", "actions": [{"name": "Terminate", "arguments": {"answer": "C"}}]}

# USER REQUEST #:
 What is x in the image?
# RESPONSE #:
{"thought": "To get the result of the equation, I need to first extract the equation from the image.", "actions": [{"name": "OCR", "arguments": {"image": "image-0"}}]}
OBSERVATION:
{"text": "x-2^3=0"}
{"thought": "The math equation is 'x-2^3=0', and I need to find x. I can solve it with a math-related tool.", "actions": [{"name": "SolveMathEquation", "arguments": {"query": "x-2^3=0, what is x?"}}]}
OBSERVATION:
{"result": "x = 8"}
{"thought": "As suggested in the last observation, the answer is 8.", "actions": [{"name": "Terminate", "arguments": {"answer": "8"}}]}

[END OF EXAMPLES]
\end{lstlisting}
\twocolumn

\onecolumn
\begin{lstlisting}[language=Python, caption=Thought templates for each action, label=lst:thought-templates]
def GetObjects_template():
    thought_templates = ["I need to check what objects are present in the {image_kw}.", 
                        "I need to analyze the {image_kw} for context."
                        "I need to identify the objects in the {image_kw}.",
                        "To answer the question, let's first analyze the {image_kw}.",
                        "To answer the question, analyzing the objects in the {image_kw} is necessary."]
    return thought_templates

def LocalizeObjects_template():
    thought_templates = ["I need to analyze the positions of {objects} in the {image_kw}.", 
                        "I need to analyze the locations of {objects} in the {image_kw}.", 
                        "I need to localize the {objects} based on the {image_kw}.",
                        "I'll identify the positions of {objects} in the {image_kw}.",
                        "I need to determine the positions of {objects} by analyzing the {image_kw}."]
    return thought_templates

def EstimateObjectDepth_template():
    thought_templates = ["I should estimate the depth of {object} to determine whether it is closer or farther.", 
                         "I will estimate the depth of {object}.", 
                         "I need to estimate the depth for {object} to make a comparison.", 
                         "To determine how far {object} is, I need to evaluate the distance to it.",
                         "I now need to estimate the depth for {object}."]
    return thought_templates


def EstimateRegionDepth_template():
    thought_templates = ["I should estimate the objects' depths to determine which one is closer.", 
                         "I need to estimate the region's depth in the image.", 
                         "I need to determine the depths of the detected objects based on their positions.",
                         "I need to estimate the depth of the objects to make a comparison.",
                         "To determine the relative proximity of the objects in the image, I need to estimate the depth of each object."]
    return thought_templates

def Terminate_template():
    thought_templates = ["Based on the information above, I can conclude that the answer is {answer}",
                         "Based on a close analysis of the {image_kw} and additional information above, I believe the answer is {answer}.",
                         "I have analyzed the {image_kw} and the information above, and I believe the answer is {answer}.",
                         "The {image_kw} and the information above suggest that the answer is {answer}.", 
                         "According to the content of the {image_kw} and the observations, I can conclude that the answer is {answer}."]
    return thought_templates
\end{lstlisting}
\twocolumn

\onecolumn
\noindent\begin{lstlisting}[caption=LLM judge prompt for MMVet, label=lst:mmvet-judge]
Compare the ground truth and prediction from AI models, to give a correctness score for the prediction. <AND> in the ground truth means it is totally right only when all elements in the ground truth are present in the prediction, and <OR> means it is totally right when any one element in the ground truth is present in the prediction. The correctness score is 0.0 (totally wrong), 0.1, 0.2, 0.3, 0.4, 0.5, 0.6, 0.7, 0.8, 0.9, or 1.0 (totally right). Just complete the last space of the correctness score.
Question | Ground truth | Prediction | Correctness
--- | --- | --- | ---
What is x in the equation? | -1 <AND> -5 | x = 3 | 0.0
What is x in the equation? | -1 <AND> -5 | x = -1 | 0.5
What is x in the equation? | -1 <AND> -5 | x = -5 | 0.5
What is x in the equation? | -1 <AND> -5 | x = -5 or 5 | 0.5
What is x in the equation? | -1 <AND> -5 | x = -1 or x = -5 | 1.0
Can you explain this meme? | This meme is poking fun at the fact that the names of the countries Iceland and Greenland are misleading. Despite its name, Iceland is known for its beautiful green landscapes, while Greenland is mostly covered in ice and snow. The meme is saying that the person has trust issues because the names of these countries do not accurately represent their landscapes. | The meme talks about Iceland and Greenland. It's pointing out that despite their names, Iceland is not very icy and Greenland isn't very green. | 0.4
Can you explain this meme? | This meme is poking fun at the fact that the names of the countries Iceland and Greenland are misleading. Despite its name, Iceland is known for its beautiful green landscapes, while Greenland is mostly covered in ice and snow. The meme is saying that the person has trust issues because the names of these countries do not accurately represent their landscapes. | The meme is using humor to point out the misleading nature of Iceland's and Greenland's names.
Iceland, despite its name, has lush green landscapes while Greenland is mostly covered in ice and snow. The text 'This is why I have trust issues' is a playful way to suggest that these contradictions can lead to distrust or confusion. The humor in this meme is derived from the unexpected contrast between the names of the countries and their actual physical characteristics. | 1.0
\end{lstlisting}

\noindent\begin{lstlisting}[caption=LLM judge prompt for MathVista, label=lst:mathvista-judge]
Please read the following example. Then extract the answer from the model response and type it at the end of the prompt.

Hint: Please answer the question requiring an integer answer and provide the final value, e.g., 1, 2, 3, at the end.
Question: Which number is missing?
Model response: The number missing in the sequence is 14.
Extracted answer: 14

Hint: Please answer the question requiring a floating-point number with one decimal place and provide the final value, e.g., 1.2, 1.3, 1.4, at the end.
Question: What is the fraction of females facing the camera?
Model response: The fraction of females facing the camera is 0.6,
which means that six out of ten females in the group are facing the camera.
Extracted answer: 0.6

Hint: Please answer the question requiring a floating-point number with two decimal places and provide the final value, e.g., 1.23, 1.34, 1.45, at the end.
Question: How much money does Luca need to buy a sour apple candy and a butter-scotch candy? (Unit: $)
Model response: Luca needs $1.45 to buy a sour apple candy and a butterscotch candy.
Extracted answer: 1.45

Hint: Please answer the question requiring a Python list as an answer and provide the final list, e.g., [1, 2, 3], [1.2, 1.3, 1.4], at the end.
Question: Between which two years does the line graph saw its maximum peak?
Model response: The line graph saw its maximum peak between 2007 and 2008.
Extracted answer: [2007, 2008]

Hint: Please answer the question and provide the correct option letter, e.g., A, B, C, D, at the end.
Question: What fraction of the shape is blue?
Choices: (A) 3/11 (B) 8/11 (C) 6/11 (D) 3/5
Model response: The correct answer is (B) 8/11.
Extracted answer: B
\end{lstlisting}

\twocolumn

\end{document}